\documentclass[10pt,twocolumn,letterpaper]{article}

\usepackage{cvpr}              

\usepackage{graphicx}
\usepackage{amsmath}
\usepackage{amssymb}
\usepackage{booktabs}

\usepackage[pagebackref,breaklinks,colorlinks]{hyperref}

\usepackage{wrapfig}
\usepackage{multirow}
\usepackage{booktabs}       
\usepackage{makecell}
\usepackage{comment}
\usepackage{mathtools, cuted}
\usepackage{amsfonts}       

\usepackage[accsupp]{axessibility}  

\usepackage[capitalize]{cleveref}
\crefname{section}{Sec.}{Secs.}
\Crefname{section}{Section}{Sections}
\Crefname{table}{Table}{Tables}
\crefname{table}{Tab.}{Tabs.}

\usepackage{soul}
\sethlcolor{SpringGreen}

\newcommand{\VLPt}[0]{VLP$^\mathcal{T}$} 
\newcommand{\VLPi}[0]{VLP$^\mathcal{I}$} 

\newcommand{\data}{\textsc{WebQA}}
\newcommand{\pmn}[1]{\begin{footnotesize}$\pm#1$\end{footnotesize}}

\newcommand{\update}[1]{\textcolor{black}{#1}} 

\def\cvprPaperID{6960} 
\def\confName{CVPR}
\def\confYear{2022}

\usepackage{overpic}
\usepackage{enumitem} 
\usepackage{overpic} 
\usepackage{color}

\definecolor{turquoise}{cmyk}{0.65,0,0.1,0.3}
\definecolor{purple}{rgb}{0.65,0,0.65}
\definecolor{dark_green}{rgb}{0, 0.5, 0}
\definecolor{orange}{rgb}{0.8, 0.6, 0.2}
\definecolor{red}{rgb}{0.8, 0.2, 0.2}
\definecolor{darkred}{rgb}{0.6, 0.1, 0.05}
\definecolor{blueish}{rgb}{0.0, 0.3, .6}
\definecolor{light_gray}{rgb}{0.7, 0.7, .7}
\definecolor{pink}{rgb}{1, 0, 1}
\definecolor{greyblue}{rgb}{0.25, 0.25, 1}

\usepackage{blindtext}

\renewcommand{\paragraph}[1]{\vspace{1em}\noindent\textbf{#1}.}
\begin{document}
\title{WebQA: Multihop and Multimodal QA}

\author{
\begin{tabular}{c@{\hspace{2em}}c@{\hspace{2em}}c}
\textbf{Yingshan Chang}$^1$ &  \textbf{Mridu Narang}$^2$ & \textbf{Hisami Suzuki}$^2$ \\
\textbf{Guihong Cao}$^2$ & \textbf{Jianfeng Gao$^3$} &  \textbf{Yonatan Bisk$^{1,3}$}\\ 
\normalfont $^1$Carnegie Mellon University & \normalfont $^2$Microsoft, Bing Search & \normalfont $^3$Microsoft Research\\
\end{tabular}
}

\maketitle
\begin{abstract}
Scaling Visual Question Answering (VQA) to the open-domain and multi-hop nature of web searches, requires fundamental advances in visual representation learning, knowledge aggregation, and language generation.  In this work, we introduce \data{}, a challenging new benchmark that proves difficult for large-scale state-of-the-art models which lack language groundable visual representations for novel objects and the ability to reason, yet trivial for humans.  \data{} mirrors the way humans use the web: 1) Ask a question, 2) Choose sources to aggregate, and 3) Produce a fluent language response.  This is the behavior we should be expecting from IoT devices and digital assistants. 
Existing work prefers to assume that a model can \textit{either} reason about knowledge in images \textbf{or} in text.  \data{} includes a secondary text-only QA task to ensure improved visual performance does not come at the cost of language understanding.
Our challenge for the community is to create unified multimodal reasoning models that answer questions regardless of the source modality, moving us closer to digital assistants that not only query language knowledge, but also the richer visual online world.
\end{abstract}
\let\thefootnote\relax\footnote{\href{webqna.github.io}{https://webqna.github.io}}

\vspace{-10pt}
\section{Introduction}

Web search is a multimodal experience: Will I find my answer on the image search tab or within text snippets? 
In contrast, most deployed Question Answering (QA) systems treat the web as a text-only landscape of facts to be extracted, ignoring the knowledge present in images.  This has two fundamental limitations: 1. The text-based web is impoverished \cite{Bisk2020,bender2020climbing}, and 2. This form of information extraction is inefficient.  For example, when searching to see if a park has picnic tables, surfacing an image of the picnic area answers the question immediately, rather than wading through pages of reviews hoping someone happened to mention this fact.  QA engines need to move to treating the Internet as a multimodal trove of information, but this requires multihop reasoning on either images or text. 

\begin{figure}[t]
    \centering
    \includegraphics[width=\linewidth]{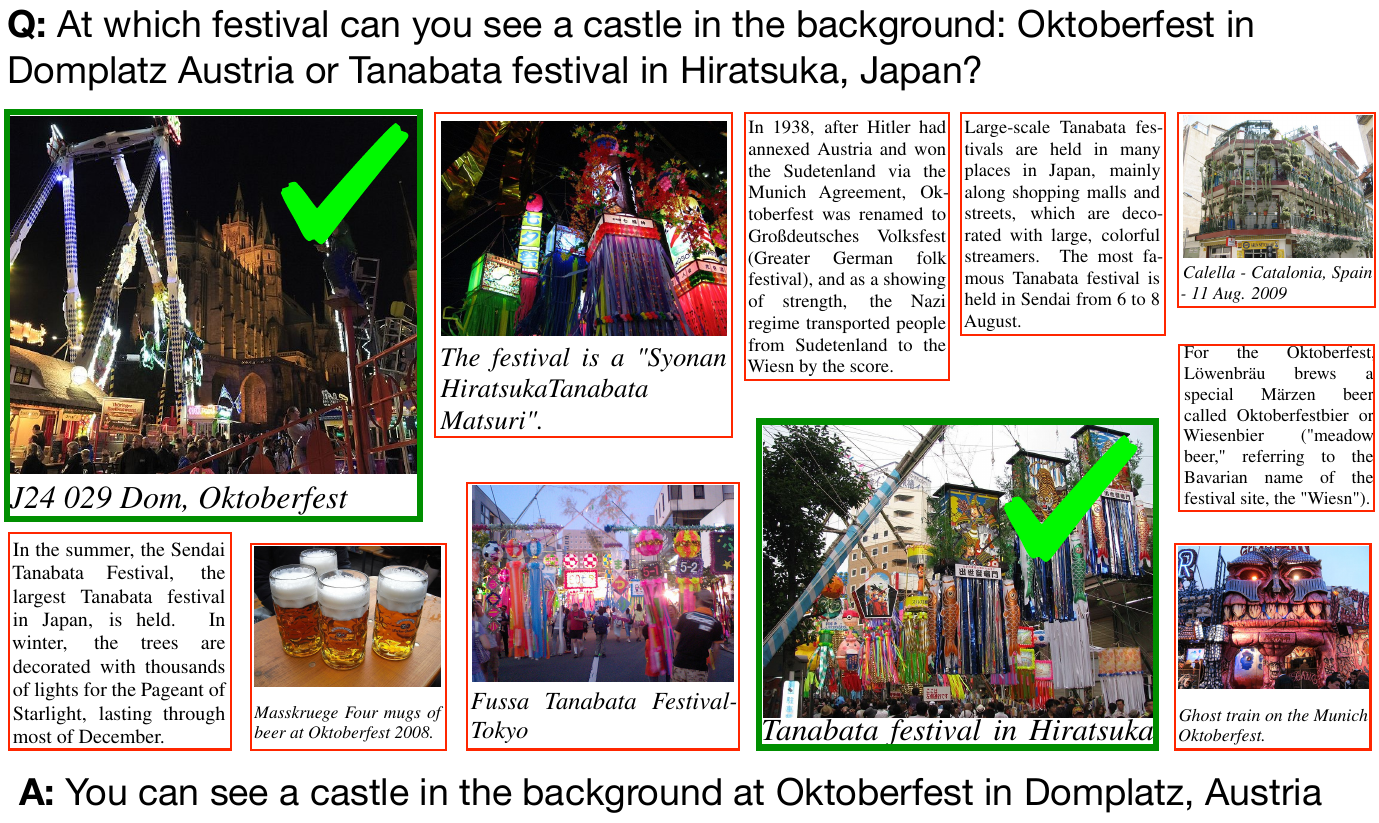}
    \caption{Example \data{} dataset pipeline in which the question requires finding and reasoning about two relevant sources and discarding distractors to produce the correct natural language answer.}
    \vspace{-10pt}
    \label{fig:teaser}
\end{figure}

To this end, datasets are rapidly emerging\cite{talmor2021multimodalqa, singh2021mimoqa,hannan2020manymodalqa}. But they either use pre-defined templates for the curation of multihop multimodal QA pairs \cite{talmor2021multimodalqa}, or encourage a ``question decomposition + rerouting to uni-modal model'' approach to superficially solve the problem\cite{hannan2020manymodalqa}. However, when humans absorb knowledge, there is no need to distinguish whether the knowledge was learned from books versus images, or whether a piece of knowledge is a composite of multiple scattered fragments versus being carried by a single one. We argue that genuine progress in reasoning over linguistic notions of meanings and visually grounded meanings under the same representation framework depends on the development of a unified system that indiscriminately treats snippets and images as knowledge carriers. On top of that, the goal includes better extraction, integration and summarization abilities in a heterogeneous information landscape.

To facilitate this research intersection, in this work we propose a novel benchmark, \data, for \textit{multi-hop, multi-modal, open-domain question-answering} where all questions are knowledge-seeking and resemble real-world use cases. Success on \data{} requires a system to a) incorporate both text and images, b) retrieve relevant knowledge in either modality, c) aggregate information from multiple sources via logical or numerical reasoning, and d) generate answers in natural language.  We experiment with state-of-the-art multimodal reasoning and text generation models, whose failures indicate promising future directions.


\begin{table}[t]
\centering
\footnotesize
\begin{tabular}{@{}l@{\hspace{4pt}}r@{\hspace{10pt}}r@{\hspace{10pt}}r@{\hspace{3pt}}r@{\hspace{10pt}}r@{\hspace{6pt}}r@{}}
 & \#Train & \#Dev & \#Test & \#Img & Len Q & Len A \\
\toprule
VQA v2 \cite{balanced_vqa_v2} & 443K  & 214K  & 453K & 200K & 6.1 & 1.2 \\
\midrule
OKVQA\cite{Marino19okvqa}                & 9.0K  & 0     & 5.0K & 14.0K           & 8.1  & 1.3 \\
MultiModalQA\cite{talmor2021multimodalqa}& 23.8K & 2.4K  & 3.6K & 57.7K           & 18.2 & 2.1 \\
ManyModalQA\cite{hannan2020manymodalqa}  & 2.0K  & 3.0K  & 5.1K & 2.9K            & --    & 1.0 \\
MIMOQA\cite{singh2021mimoqa}             & 52.4K & 0.7K  & 3.5K & 400.0K & --    & -- \\
\midrule
\midrule
\data{} (ours) & 34.2K & 5K & 7.5K & 390.0K & 17.5 & 12.5 \\
\bottomrule
\end{tabular}
\caption{Comparison of multimodal knowledge-seeking benchmarks by size and average question/answer lengths.\protect\footnotemark }
\label{tb:Comparison1}
\vspace{-5pt}
\end{table}

\section{Related Work}
\footnotetext{Note, MultiModalQA and ManyModal QA also contain tables -- 3.5K for ManyModal and while 700k were used MultiModalQA's dataset generation, it is unclear how many ended up in the final dataset.}
Many datasets and tasks can be broadly considered ``question answering.'' For example, VQA \cite{balanced_vqa_v2, VQA, Marino19okvqa, Hudson19gqa} is one of the widely studied tasks at the intersection of language and vision. Nevertheless, it is unclear how VQA models should be adapted to open-domain scenarios. This is largely due to the simplification of VQA tasks into classification over a fixed vocabulary of frequent answers. Recent work on video \cite{yagcioglu2018recipeqa, lei2018tvqa, tapaswi2016movieqa} has also adopted a multiple-choice format. 
In contrast, OK-VQA \cite{Marino19okvqa} broadens the task to knowledge-seeking questions. OK-VQA and our task differ in the role of images. Images in OK-VQA are regarded as part of the query rather than as part of the knowledge source, and can only be processed after retrieval.  

Within the natural language community, QA datasets are experiencing a similar transition from multiple-choice and span prediction to the harder free-form answer generation paradigm. Multi-hop question answering has recently taken the spotlight as it aligns with the multi-hop nature of how humans perform reasoning during knowledge acquisition, leading to a proliferation of benchmarks \cite{welbl2018constructing, Yang18hotpot, talmor2018web}.

There have been several recent benchmarks for reasoning over input and contexts in multiple modalities \cite{Suhr2019:nlvr2}. MultiModalQA \cite{talmor2021multimodalqa} made the first foray into complex questions that require reasoning over snippets, tables and images. It focuses on 
cross-modal heterogeneous knowledge extraction. However, questions 
are generated from 
templates. Once a template is detected the task reduces to filling in blanks with modality-specific answering mechanisms. 

ManyModalQA \cite{hannan2020manymodalqa} also deals with snippets, images and tables. However, the primary challenge their design addresses is the choice of answer modality --- rather than knowledge aggregation or extraction.  
Our focus is more about representing world knowledge in a unified space, than about distinguishing the answer modality, since mastering the former may naturally eliminate the need to classify questions according to the answer modality.

Finally, MIMOQA \cite{singh2021mimoqa} introduces a new concept of ``Multimodal Input Multimodal Output'' which highlights accompanying a textual answer with an image in order to enhance cognitive understanding. MIMOQA requires selecting a text span and an image from the context as an 
output pair. Their approach is nicely complementary to ours. 
Where we differ, is that our task also requires aggregation and summarization before producing the final natural language answer, whereas the outputs required by MIMOQA are not completely digested by the model. Here, we refer ``digesting" to the ability to produce a reasonable output which cannot be directly copied from the input.

\begin{table}[t]
\footnotesize
\begin{tabular}{@{}l@{\hspace{5pt}}l@{\hspace{5pt}}l@{}}
    & Eval Metrics & Answer Schema \\
    \toprule
VQA v2 & \multirow{2}{*}{$min\{\frac{\#\mathrm{human\  agreement}}{3}, 1\}$} & \multirow{2}{*}{Top training answers}\\
OK-VQA & & \\ 
\midrule
MultimodalQA & \makecell[l]{Exact Match\\ F1} & \makecell[l]{Txt: span/Y/N \\ Img: Fixed vocab \\ Table: Y/N, cell, or op.} \\
\midrule
ManymodalQA & Classification Accuracy & \makecell[l]{Context word or vocab} \\
\midrule
MIMOQA & \makecell[l]{Txt: ROUGE-1/-2/-L or BLEU \\
        Img: Precision@1/@2/@3} &
        \makecell[l]{Span prediction\\ + Image retrieval} \\
\midrule
\midrule
\data{} (ours) & \makecell[l]{Fluency:\phantom{Acc: } BARTScore \\ Keyword Acc: Recall/F1} & Complete NL sentence \\
\bottomrule
    \end{tabular}
    \caption{Comparison of knowledge-seeking, multimodal benchmark  metrics and answer schema.}
    \label{tb:Comparison2}
\vspace{-10pt}
\end{table}

Tables \ref{tb:Comparison1}, \ref{tb:Comparison2} and Appendix E provide comparisons between \data{} and related datasets. 
No existing multimodal or knowledge-seeking benchmark requires the answers to be complete, free-form natural language sentences, as opposed to extractive spans, or elements from a finite set. Additionally, previous work has not supported both natural language generation (NLG) evaluation and accuracy-style evaluation as we do. To this end, we highlight that \update{a) in \data{} more importance is attached to digesting, aggregating and summarizing information as answers cannot be simply copied from an existing text span or image patch, b) \data{} requires the source retrieval stage in addition to VQA, which better simulates the full reasoning pipeline during a web search}, and c) answers in the form of a natural language sentence better transit to downstream applications such as conversational agents and voice assistants. 
\section{Task Formulation}
As in Fig~\ref{fig:teaser}, examples consist of a question $Q$, a set of positive sources $s_1, ..., s_m$ (in green), a set of distractor sources $s_{m+1}, ..., s_n$ (in red) and an answer $A$. Each source can be either a snippet or an (image, description) pair. Each image is accompanied by a description to resolve names or geographic information not present in the image itself, but serve as critical links to references in the question. \update{We include both a restricted ($n\!\approx\!40$) and full ($n\!\approx\!900K$) setting.}

We decompose the task into two stages. First, given $Q$ and $s_1, s_2, ..., s_n$, the model identifies the sources from which to derive the answer. The second stage is question answering where the model takes $Q$ and the chosen sources as context $C$, to generate an answer $A$.  Ideally, a single-stage system would jointly process $Q, s_1, s_2,\ ...,\ s_n$ to produce $A, C$, but we are unaware of any modeling approaches that can consume sufficiently large multimodal contexts to achieve this, so this is left to future work.

\section{\data{}}

Following the paradigm popularized by search engines, we structure our data as having answers that can be found either via image search or general web (text) search. \update{Note, WebQA does not contain questions that need an image and an (independent) snippet as knowledge sources. However, \textit{all} image-based questions already require processing \textit{both} images and text as \textit{image descriptions} provide necessary information.} Below we outline how both types of questions are collected, structured, and filtered for quality.

\subsection{Answers from Images}

We collect both multi-image questions that require stitching two images to answer and complex single-image questions. Rich multi-image questions do not naturally exist at scale in user search logs,\footnote{While details are omitted here, we requested details from a search company that provided us basic statistics about query logs to confirm this.} likely because users do not issue queries they believe search engines cannot handle, thus we turn to crowdsourcing.

We presented annotators with a set of six related images and asked them to produce three QA-pairs by selecting one or two images per pair that are necessary to answer the question.
We require that at least one of the three pairs utilizes two distinct images.
Additionally, we instructed annotators to avoid questions that: a) are simple facts (e.g. ``\textit{How many wheels does a car have}"); b) are easily answered by a text-only search; c) are bound to a specific image; d) ensure every question is meaningful without paired context. This elucidates one of the key differences between the well-known VQA task and ours. In most VQA style tasks, every question is about a paired image, whereas in our task images serve as knowledge sources over which to reason, and do not serve the role of augmenting the question. To assist annotators, each image is accompanied by a description extracted from Wikipedia. This description is only to be used to confirm the name or location of the objects depicted. The answer has to be derived from visual clues.

Images were crawled from Wikimedia Commons via the Bing Visual Search API. \update{Wikimedia's topic list cannot be used directly as most categories are (visually) uninteresting. We seeded with natural scenes and iteratively refined the image pool by removing categories flagged as (visually) uninteresting. This resulted in categories like animals, plants, attractions, and architecture (Fig~\ref{fig:topic_cloud}). 
}

\begin{wrapfigure}[13]{r}{0.5\linewidth}
    \vspace{-15pt}
    \hspace*{-10pt}\includegraphics[scale=0.13]{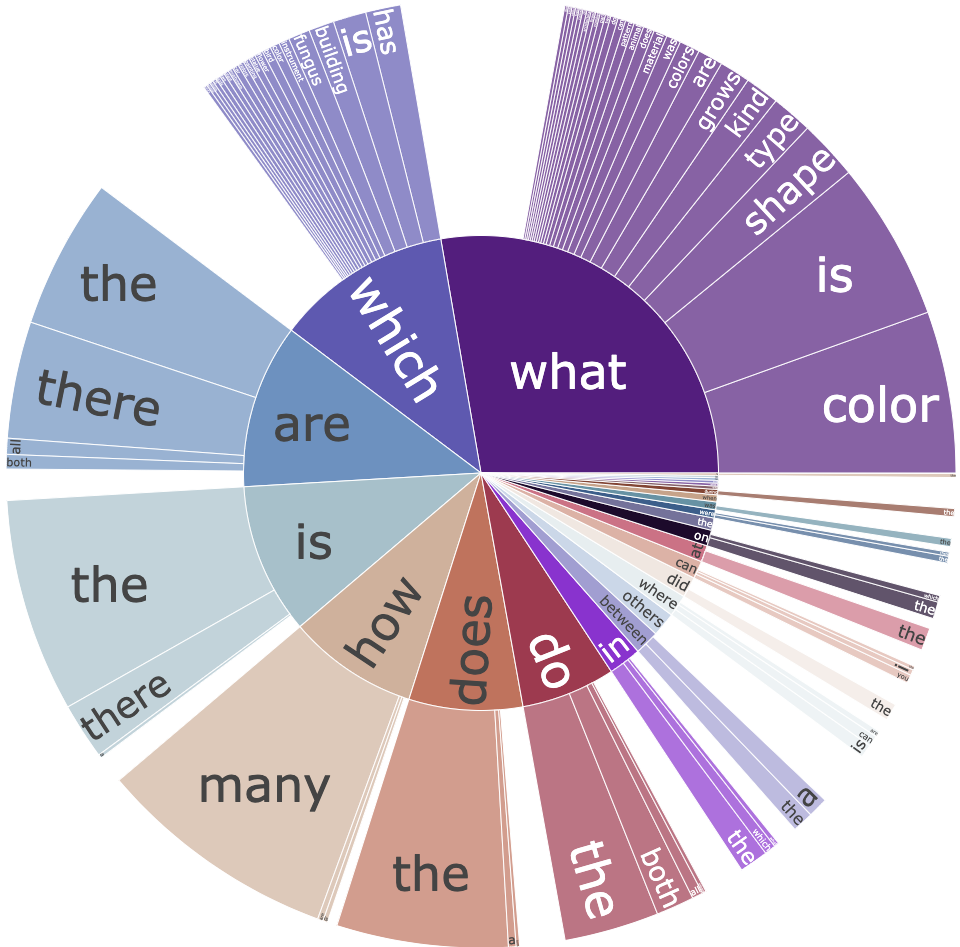}
    \vspace{-5pt}
    \caption{Image question prefixes (see Appendix B).}
    \label{fig:plotly_img_data}
\end{wrapfigure}

\paragraph{Hard Negative Mining}
We produce a set of both text- and image-based hard negatives for models to sift through for every question. Text sources are extracted from relevant passages on Wikipedia based on noun chunks in the question, while limiting overlap to avoid false negatives.
For images, we leverage Bing APIs to find similar images with respect to both the description (via Bing Image Search) and the visual content (via Bing Image Insight).
In total, we collect 25K image-based questions, each requiring an average of 1.4 visual sources, and paired with 15.3 text and 15.9 visual distractors. Question prefixes are visualised in Fig~\ref{fig:plotly_img_data}.

\paragraph{Categorization} We categorize questions into open and closed classes.  Closed class questions include: color, shape, number (i.e.``how many''), yes/no (Y/N), and ``multi-choice'' (MC). The rest are open class questions.

\paragraph{Adversarial splits} We construct our test set to be out-of-distribution when possible to reward models with better generalization and reasoning.
For color, shape, and number questions, we partition the answer set and ensure that the majority class during training does not carry over to testing. For the ``Y/N" and ``MC" classes, we trained models on 10 random train-test splits and consistently difficult samples across splits were placed in the test set. Finally, we randomly split questions from the open-class ``other''.

\subsection{Answers From Text}
We collected multi-hop QA pairs that involve combining knowledge from $\geq$2 snippets. To generate diverse, yet consistent, topics for mining difficult multi-hop reasoning questions, we construct clusters of similar entities, but where text snippets had low overall n-gram overlap or semantic similarity (yielding 8K clusters).
We provide annotators with four snippets to prevent 
and 
allow them to contribute facts they researched to help answer the question.

\paragraph{Hard Negative Mining}
For text distractors we mine passages from Wikipedia that contain noun phrases from the question and choose those with the highest lexical overlap but lacking reference to the answer. For image distractors, we use the images and descriptions present on the aforementioned Wikipedia pages, again filtering for those with high lexical overlap.
In total, we collected 24K text-based questions, each requiring 2.0 text sources, and paired with 14.6 text and 11.6 visual distractors. Lacking clear criteria for question categorization, we do not construct an adversarial test split, but instead simply sample randomly.

\begin{table}[t]
\centering
\small
\begin{tabular}{@{}l@{\hspace{4pt}}c@{\hspace{4pt}}c
                    @{\hspace{4pt}}c@{\hspace{4pt}}c
                    @{\hspace{4pt}}c@{\hspace{4pt}}c@{}}
\toprule
         &          &        & \multicolumn{2}{@{\hspace{5pt}}c}{Descriptions}& \multicolumn{2}{c}{Snippets}\\
      & Question & Answer & Correct & Distract & Correct & Distract \\
\midrule
Image & 16.4\pmn{\phantom{0}6} & 14.4\pmn{\phantom{0}6} & 13.3\pmn{11} & 12.6\pmn{11}  & ---            & 36.4\pmn{10} \\
Text  & 18.6\pmn{\phantom{0}8} & 10.7\pmn{10}& ---            & 14.1\pmn{13} & 45.3\pmn{12} & 38.3\pmn{10} \\
\bottomrule
\end{tabular}
\caption{Length distribution for different textual components.  
}
\label{tb:length_distr}
\end{table}

\subsection{Quality Control}
We ensure the data quality via crowdworkers training and expert-feedback-in-the-loop, which are found to be effective ingredients in crowdsourcing \cite{nangia2021ingredients}. 
The initial pool of annotators were trained with a tutorial and selected via a qualification task. 
Additionally, we released the annotation task in batches to spot check quality after every batch, followed by sending constructive feedback to correct any deviation from our expectations. Workers who failed multiple times were de-qualified. Crowdsourcing data is challenging in that crowdworders are usually income-driven and will stick to a fixed answer generation pattern once they find it lucrative. To better align the crowdworkers' incentives with our goal, we generously bonus out-of-the-box thinking. All data was then also run through additional validation HITs to ensure agreement. Annotator pay averaged \$13/hr overall (lower on the initial qualification and higher on the annotation/validation). \update{Appendix A contains rubics and interfaces.}

\subsection{Dataset Statistics}

\newlength{\oldcolumnsep}
\setlength{\oldcolumnsep}{\columnsep}
\setlength\columnsep{10pt}
\newlength{\oldintextsep}
\setlength{\oldintextsep}{\intextsep}
\setlength\intextsep{0pt}

\begin{wraptable}[7]{r}{0pt}
\small
\begin{tabular}{@{}l@{\hspace{7pt}}c@{\hspace{7pt}}c@{\hspace{7pt}}c@{}}
       
\toprule
Modality & Train & Dev & Test \\
\midrule
Image & 18,954 & 2,511 & 3,464  \\
Text  & 17,812 & 2,455  & 4,076 \\
\bottomrule

\end{tabular}
\caption{Number of samples collected for each modality fold.}
\label{tb:num_of_samples}
\end{wraptable}

In total, \data{} has over 34K training QA pairs, with an additional 5K and 7.5K held out for development and testing. Overall Statistics are summarized in Table \ref{tb:num_of_samples} and language distributions are presented in Table \ref{tb:length_distr}.\\

\setlength{\columnsep}{\oldcolumnsep}
\setlength{\intextsep}{\oldintextsep}

\begin{figure}[t]
\centering
  \begin{subfigure}[b]{0.48\linewidth}
    \includegraphics[width=0.98\textwidth]{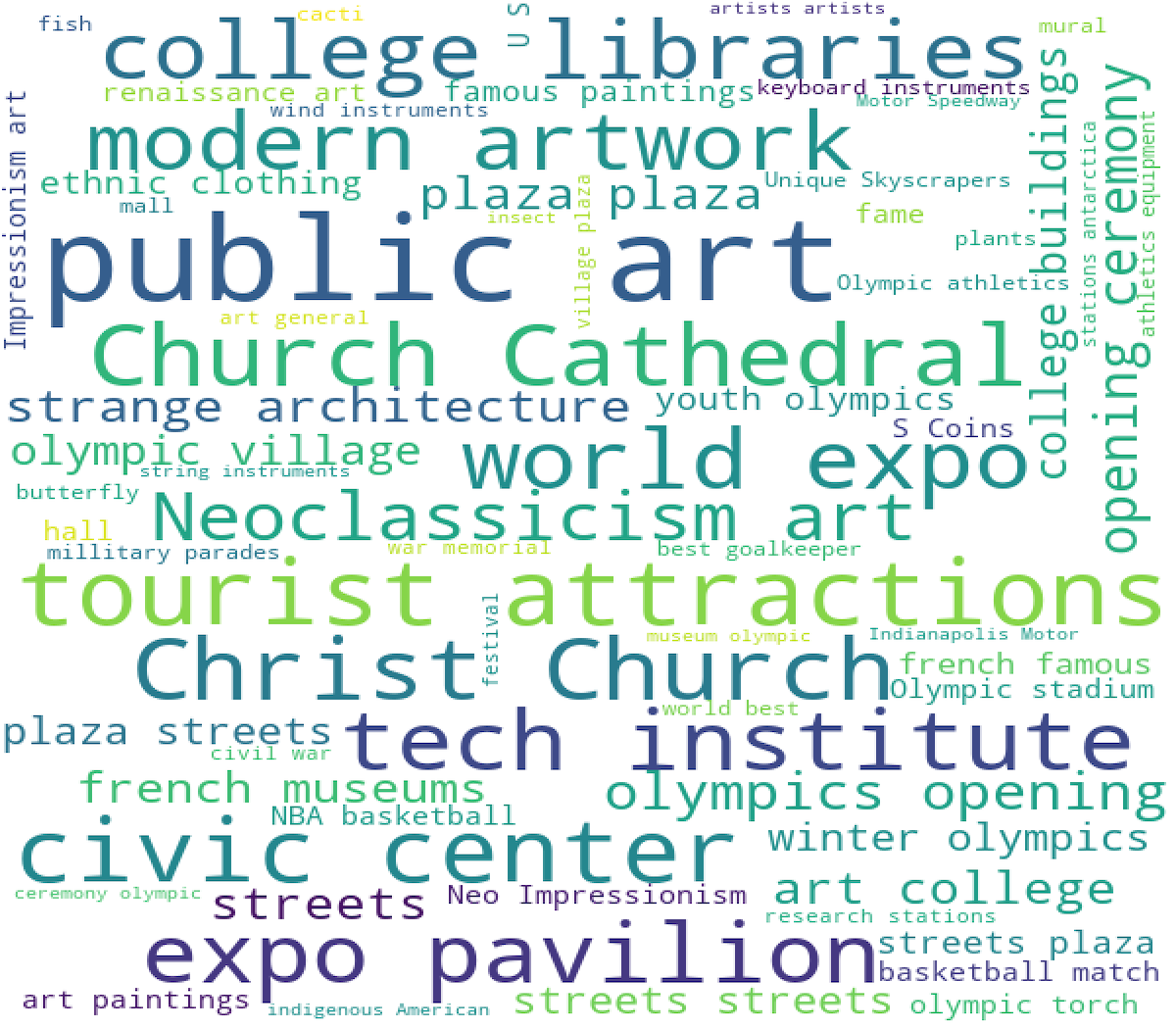}
    \label{fig:img_topic_cloud}
  \end{subfigure}
 ~
  \begin{subfigure}[b]{0.48\linewidth}
    \includegraphics[width=0.98\textwidth]{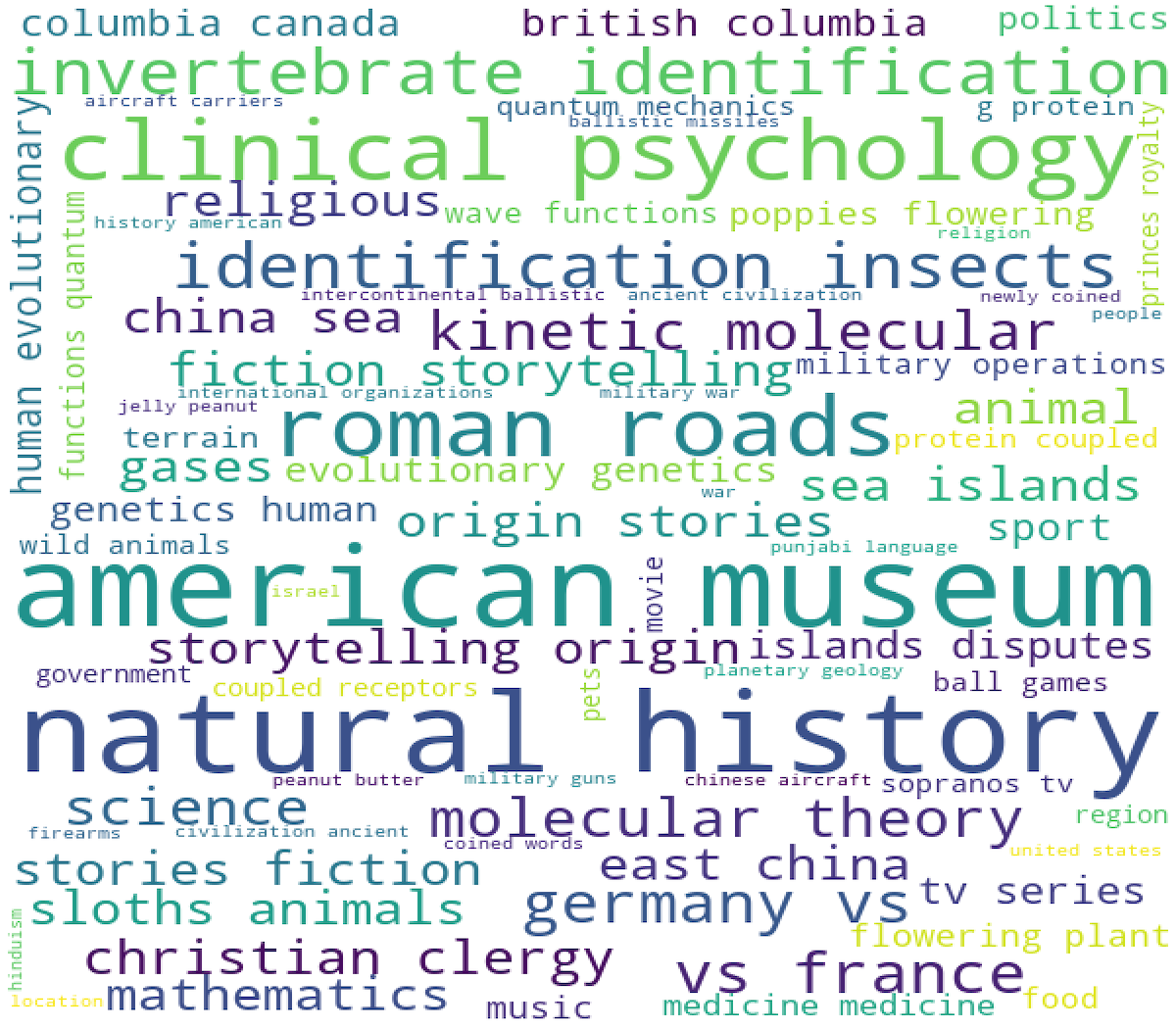}
    \label{fig:txt_topic_cloud}
  \end{subfigure}
  \vspace{-5pt}
  \caption{Samples of common topics in the image-based (left) and text-based (right) folds of the data.  
  }
\label{fig:topic_cloud}
\end{figure}
\vspace{-15pt}

\paragraph{Multi-hop} 44\% of image-based queries and 99\% of text-based queries require two or more knowledge sources. This is verified by crowdworkers during validation to ensure that multiple knowledge sources provide non-overlapping information and cannot be replaced by each other. Additionally, as image sources also require understanding the caption, even single-image queries require multi-source reasoning.

\paragraph{Topics} Fig \ref{fig:topic_cloud} provides a qualitative sense of the wide range of topics covered in \data{}. 
In contrast to MultiModalQA, the images in \data{} concentrate on the natural-world, events, and locations rather than digital artifacts (e.g. posters/logos). 
Snippets also exhibit a wide range of topics from contemporary science to ancient mythology. When comparing the topic clouds, it is clear that image-based queries more often relate to physical entities while text-based queries tend to be more abstract.
\section{Metrics}
\data{} requires a model to answer open-domain questions and cite its sources. Therefore, we evaluate model performance with respect to both relevant fact prediction and question answering. While fact retrieval is easily evaluated via F1, language fluency and accuracy metrics are nuanced. 

\subsection{Question Answering Metrics}
Our task expects fluent and complete sentences as answers, which we believe are appropriate for applications such as voice assistants or conversation agents. Therefore, the quality is measured as both fluency and accuracy. 
On each testing sample we collected five full-sentence answers written by humans. In addition, we collected one keyword answer by asking human annotators to rephrase the full-sentence answer into a succinct minimal semantic form. 

\paragraph{Fluency} We measure fluency via BARTScore \cite{yuan2021bartscore}, a newly proposed NLG evaluation metric based on accurate measurement of paraphrase quality. 
BARTScore($A$, $B$) measures the probability of generating $B$ from $A$. In our setting, this is computed as 
BARTScore(\textit{r}, \textit{c}), which can be interpreted as the probability of generating a candidate given a reference. 
Since BARTScore is based on the generation likelihoods, it does not distribute neatly across $[0,1]$. So we normalize BARTScore(\textit{r}, \textit{c}) by the identity score BARTScore(\textit{r}, \textit{r}). On top of that, we make the normalized score bounded by 1. Finally, we choose the best score for a candidate across all references, as illustrated in Eq.~\ref{eq:1}.

\vspace{-10pt}

\begin{equation} 
\label{eq:1}
\mathbf{FL}(c, R) = max\big\{ min\big(1, \frac{BARTScore(r, c)}{BARTScore(r, r)}\big) \big\}_{r\in R}
\end{equation}

This formulation
a) prioritizes semantic agreement and is robust to functional words misplacement, b) does not heavily punish short sentences (i.e. $<$ 4 words) as BLEU4 \cite{papineni2002bleu} does, 
c) 
penalizes word reordering / disfluencies 
d) and unlike BERTScore \cite{zhang2019bertscore}, which indiscriminately treats all colors or all shapes as nearly identical, BARTScore better captures small but critical differences. However, no language based embedding metrics accurately evaluate visual phenomena, so we also introduce an accuracy metric.

\paragraph{Accuracy} 
To ensure answer accuracy we use the collected keywords.  Note, our paradigm differs from both open-domain text QA which focuses on lexical F1 and visual QA which uses a multiple choice evaluation.  F1 rewards copying the question 
even if the key information is missing (e.g. the wrong color or count is chosen).  Conversely, multiple-choice paradigms are not applicable to evaluate generated sentences. The goals of measuring accuracy on \data{} are: 
1. Detect the presence of key entities.
2. Penalize the use of any incorrect entities. 
3. Avoid penalizing semantically relevant but superfluous words.
We are unaware of any solution to all of these criteria in the naturally mixed setting of our data (open-domain entities with a nearly closed-domain set of properties), so we propose an appropriate metric to tackle the different styles of answers.

\setlength{\oldcolumnsep}{\columnsep}
\setlength\columnsep{10pt}
\setlength{\oldintextsep}{\intextsep}
\setlength\intextsep{0pt}

\begin{wraptable}[  8]{r}{0.45\linewidth}
\centering
\footnotesize
\begin{tabular}{@{}l@{\hspace{5pt}}l@{}}
$qc$     & "Answer Domain" $D_{qc}$ \\
\toprule
       & Union of keywords \\
color  & ... across color queries \\
shape  & ... across shape queries \\
number & ... and \#s in references \\
\midrule
Y/N & \{'yes', 'no'\} \\
\bottomrule
\end{tabular}
\caption{``closed'' classes}
\label{tb:metric_answer_domain}
\end{wraptable}
Given the aforementioned question categorization for visual queries, questions having closed answer domains should be evaluated via F1 that tests for precision (avoiding a model producing both Yes and No to game the metric). 
We define the answer domains $D_{qc}$ of those question categories ($qc$) in Table \ref{tb:metric_answer_domain}.
For the remaining visual queries and all textual queries, they have diverse and unrestricted answer domains. So, there are good reasons to believe that the probability of cheating by guessing a long list of keywords is small and would be penalized by BARTScore, so we evaluate accuracy via recall (RE). With $c$ as a candidate output, $K$ for correct answer keywords, and $qc$ for question category, Equation \ref{eq:2} sketches our $\mathbf{Acc}$ score.

\setlength{\columnsep}{\oldcolumnsep}
\setlength{\intextsep}{\oldintextsep}

\vspace{-10pt}
\begin{equation} \label{eq:2}
\mathbf{Acc}(c, K) = \left\{
    \begin{array}{lr}
        \mathrm{if}\  qc \in \mathrm{[color, shape, number, Y/N]}: \\
        \hspace{2em} F1\big(c\cap D_{qc}, K\cap D_{qc}\big)  \\
        \mathrm{otherwise:} \\
        \hspace{2em} RE\big(c, K\big)                                 \\
    \end{array}
\right.
\end{equation}

Finally, we report the average combined fluency and accuracy score $\mathbf{FL}$*$\mathbf{Acc}$ across all test samples as a single evaluation result for a system.\footnote{
Our metric does not solve NLG evaluation. Specifically, the ``MC" question type often takes the form: "which one in set $S$ has property \textit{xyz?}". Unlike the categories in Table \ref{tb:metric_answer_domain} where it is wrong to output incorrect elements, including more elements in additional to the correct element in an answer may be correct if asked to compare the elements. We leave this to future NLG evaluation research as outside the scope of this work.
}

\section{Baseline Models}

We test existing models on \data{} in both fune-tuned and few-shot settings. The former fine-tunes a pre-trained vision-and-language transformer \cite{Zhou20VLP} on our source retrieval and QA tasks, while the latter (PICa \cite{yang2021empirical}) prompts GPT-3\cite{Brown2020} with engineered prefixes. 
Note, since the answer space in \data{} is inappropriate for the classification approach (3K answers) considered by most VQA models, these models \cite{tan2019lxmert, lu2019vilbert, chen2020uniter, su2019vl}, cannot be applied in our generative task.\footnote{see \update{Appendix C for limitations of classification in \data{}}} At present, \update{VLP \cite{Zhou20VLP} and Oscar \cite{li2020oscar} are the top generative multimodal transformers. Oscar is built on VLP so we chose VLP as more canonical but include the state-of-the-art visual features of VinVL implemented in Oscar+ \cite{zhang2021vinvl}. Other recent models \cite{cho2021unifying} may also have complementary strengths. 
} 
To test the largest possible language model, we also run PICa \cite{yang2021empirical} which leverages VinVL based captioning to augment GPT-3 \update{with oracle source knowledge}. 
\update{Finally, to simulate the full retrieval setting, we ran zero-shot sparse and dense retrieval models over the entire collection of sources.}

\subsection{Fine-tuning Approach}
We train two separate models for source retrieval and question answering on from released VLP \cite{Zhou20VLP} weights. 

\paragraph{Input Representation} 
Text segments, including the questions, answers, textual sources and image captions, are tokenized by the \texttt{Bert-base-cased} \cite{devlin2018bert} tokenizer. Each image is represented by 100 regions predicted by an object detection model, which is a variant of Faster RCNN with an ResNeXt-101 FPN backbone, pretrained on Visual Genome \cite{krishna2017visual}. We take the output of fc1 layer from the object detection network an 2048-dim feature and finetune the fc2 layer. We also experiment with the latest state-of-the-art representations from VinVL\cite{zhang2021vinvl}. Comparing to ResNeXt-101 FPN, the major advances of VinVL include a larger backbone (ResNeXt-152), replacement of FPN by C4\footnote{Prior work \cite{zhang2021vinvl, jiang2020defense} has shown that C4 features are more effective for VL tasks due to its ImageNet weight initialization and inductive bias of the convolutional head. Both factors are not present in the MLP head of FPN.} and better pretraining enriched by attribute information.

\paragraph{Source Retrieval}
Candidate sources $s_1, s_2,\ ...,\ s_n$ are fed to the model one by one. Each pass takes the concatenation of $<$\texttt{[CLS]}, $s_i$, \texttt{[SEP]}, $Q$, \texttt{[SEP]}$>$ and estimates probability of a particular source being selected. Let $\mathcal{G}$ and $\mathcal{D}$ denote the set of gold sources and distractors for a sample. The loss function is as follows.

\vspace{-10pt}
\begin{equation} \label{eq:3}
Loss_{retrieval}= \sum\limits_{s_i \in \mathcal{G}}logp_{s_i} + \sum\limits_{s_i \in \mathcal{D}}log(1-p_{s_i})
\vspace{-10pt}
\end{equation}

\paragraph{Question Answering}
 We feed $<$\texttt{[CLS]}, $S$, \texttt{[SEP]}, $Q$, $A$, \texttt{[SEP]}$>$ to the Transformer, where attention masks are applied to tokens in $A$ to satisfy the auto-regressive property. We use standard Masked-Language-Modeling  \cite{devlin2018bert} loss during fine-tuning. We decode by iteratively appending a \texttt{[MASK]} to the end of the input, replacing it with a predicted token and appending a new \texttt{[MASK]} for the next timestep. Generation stops upon seeing \texttt{[SEP]}, \texttt{[PAD]}, or reaching a maximum length. We use beam search ($n\!=\!5$) and choose the most confident output for evaluation.
 
\paragraph{Model Variants}
 In addition to the standard VLP trained on full data, we also include two modality-specific variants \VLPi{} and \VLPt, which are trained on image- or text-based queries only as opposed to the full data, in order to reveal gains and losses resulted from the complexity of presenting models with data from both modalities,

\subsection{Zero-shot Full-scale Retrieval Approach}

For end-to-end performance in an open-domain setting, we consider the entire collection of sources as our retrieval space (390k images and 540k text sources). Since running VLP-based retrieval of the test set over the entire source collection is prohibitively expensive ($\sim$3 years), \update{we consider both sparse retrieval (BM25 \cite{robertson2009probabilistic}) and dense retrieval for a coarse filtering. Dense retrieval was achieved via CLIP \cite{radford2021learning} encoding all image and text sources, as well as all questions.}\footnote{CLIP never assigns an image as more similar to a question than any text snippet, so we assume knowledge of what modality to retrieve. A BERT modality classifier can also achieve near perfect accuracy, but future unified approaches will hopefully not require this simplification.} Next, using the modality knowledge, we rank all image/text sources based on the question-source similarity.

\subsection{Few-shot Question Answering Approach}

PICa \cite{yang2021empirical} is the strongest model on OK-VQA \cite{Marino19okvqa}, where GPT-3 is prompted to generate answers given a few training samples as prefix. We adapt PICa to our QA task using oracle sources to provide an upper-bound for the best possible performance of the strongest known models. PICa (and GPT-3) exhibit unstable behaviors on source prediction when presented with $>$4 choices (as it is most familiar with 4-way multiple choice tasks). Due to the inability to fine-tune, we cannot construct a truly fair comparison of PICa with our other baselines on our full pipeline.   

We construct an input prefix by concatenating the pre-selected training examples, the context and the question of a testing sample. Since PICa's transformer backbone does not accept visual input, each image is described by three text segments, namely 1) a wikipedia description, b) a caption generated by Oscar+\cite{zhang2021vinvl} and c) a list of tags predicted by Oscar+. Limited by the maximum input length, we experimented with an 8-shot setting. If the input length exceeds the maximum length, we decrease the number of shots until it fits in the length budget.  

\begin{figure}[t]
\begin{center}
    \includegraphics[width=\linewidth]{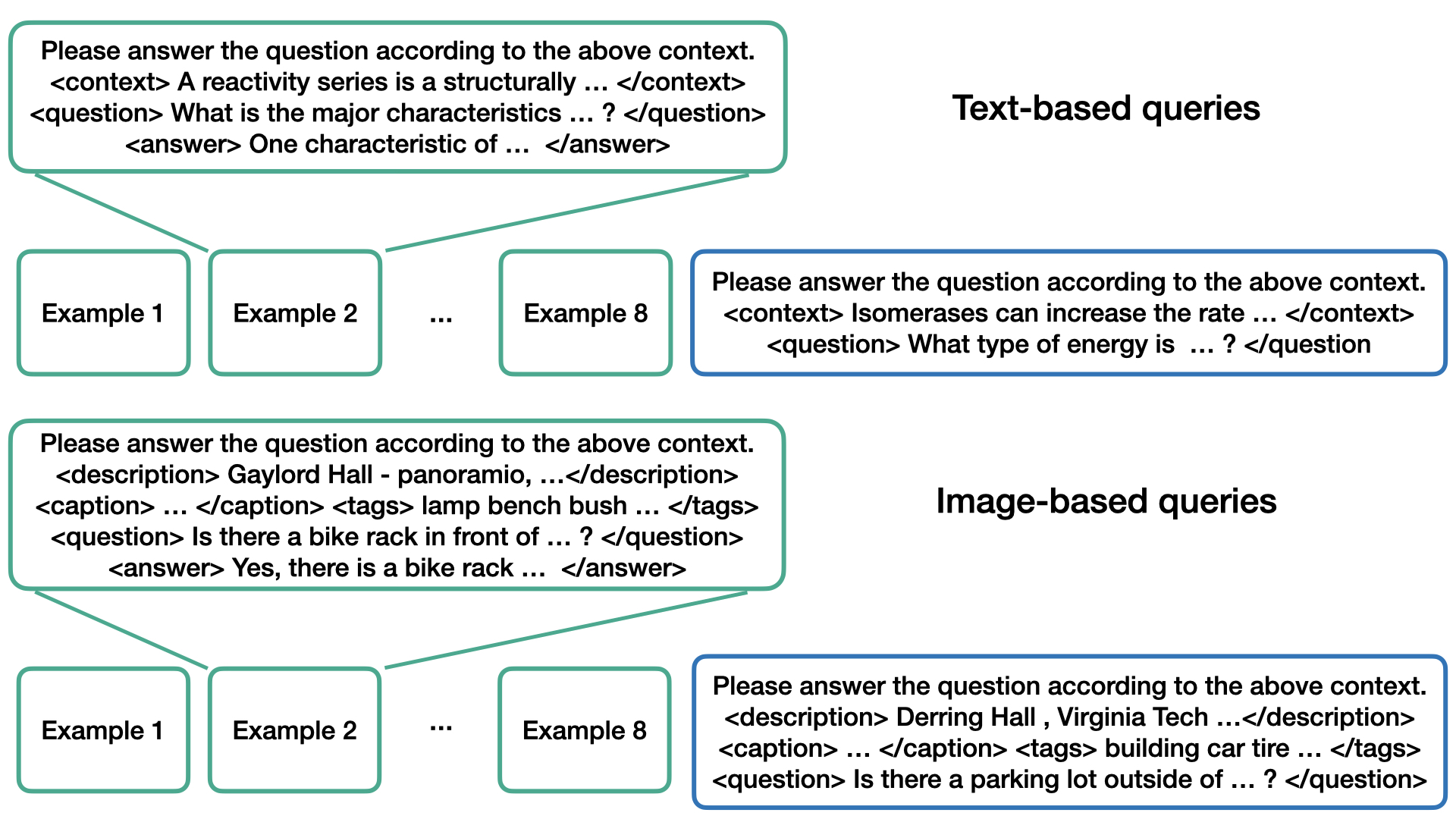}
    \caption{Few-Shot GPT-3 prompts.}
    \vspace{-20pt}
    \label{fig:prompts}
    \end{center} 
\end{figure}

\paragraph{Training Example Selection} Training examples to be included in the prefix for each testing sample are selected according to both question and source similarities. We use CLIP\cite{Radford2021} to extract text or image encodings for questions, oracle snippets and oracle images. When multiple sources exist, we take the average of pairwise similarities between sources in one sample and sources in the other.

\paragraph{Prompt Design} We use XML-style brackets \cite{jiang2021delphi} to denote different text segments. See Fig~\ref{fig:prompts} for what constitutes a prompt for a text- or image-based query.

\section{Results \& Analysis}
Below we present results and analysis of our baselines' performance on \data{}. We include question-only baselines for both VLP and PICa to investigate how effectively models use the sources. 
VLP scores 22.6 on the proposed metric when evaluated end-to-end (Table \ref{tb:Main}). Modest improvement can be achieved by knowing the gold sources, showing room for growth on retrieval correctness. We observe that the latest best-performing visual encoder, VinVL, does not lead to significant gains. This may support the argument that the missing aspects from the status quo are more reflected in cross-modal information sharing than in the imperfection of uni-modal representations. PICa achieves a large gain over VLP. Promising as it is, we later show that, while pursuing the benefits of scaling up is one thing, there is still a lot remaining to be done to combat the diminishing returns involved with scale \cite{abnar2021exploring}. We show that humans can perform our task with ease (i.e. achieving $>$94 $\mathbf{Acc}$ and $>$55 $\mathbf{FL}$) computed via cross-evaluation on multiple (3-6) references provided by different annotators to prove robustness and consensus. While models' $\mathbf{FL}$ scores are high, reaching human-level accuracy is not within sight.

\begin{table}[t]
\centering
\small
\begin{tabular}{@{}l@{\hspace{4pt}}l
                @{\hspace{2pt}}c
                @{\hspace{1em}}c@{\hspace{4pt}}c@{\hspace{4pt}}c
                @{\hspace{1em}}c@{\hspace{4pt}}c@{\hspace{4pt}}c@{}}
\toprule
 & & Source & \multicolumn{3}{@{}l@{}}{QA Pred. source} & \multicolumn{3}{@{}c@{}}{QA Oracle source}  \\
& & $\mathbf{F1}$  & $\mathbf{FL}$& $\times \mathbf{Acc}$ &$ = $ & $\mathbf{FL}$ & $\times \mathbf{Acc}$ & $= $ \\
\midrule
\multirow{6}{*}{\rotatebox{90}{\footnotesize Restricted}} & VLP (Q-only) & ----- & 34.9 & 22.2 & 13.4 & 34.9 & 22.2 & 13.4 \\
& VLP           & 68.9 & 42.6 & 36.7 & 22.6 & 44.6 & 40.4 & 24.5 \\
&\hspace{2pt} + VinVL     & \textbf{70.9} & \textbf{44.2} & \textbf{38.9} & \textbf{24.1} & 45.7 & 42.2 & 25.9 \\
\cmidrule{2-9}
& PICa (Q-only) & --- & --- & --- & --- & 47.6 & 43.4 & 28.8 \\
& PICa & --- & --- & --- & --- & 57.1 & 61.6  & 40.1 \\
\midrule
\midrule
\multirow{2}{*}{\rotatebox{90}{\footnotesize Full}} & \update{CLIP$_{(2)\phantom{0}}$+VLP}  & 12.0 & 34.2 & 24.1 & 14.6  & --- & --- & --- \\
& \update{CLIP$_{(20)}$+VLP} & \textbf{24.0} & \textbf{36.1} & \textbf{27.2}  & \textbf{16.1} & --- & --- & --- \\
\midrule
\midrule
& Human         & \textbf{90.5} & ---   & ---   &  ---  & \textbf{55.1} & \textbf{94.3} & \textbf{52.4} \\
\bottomrule
\end{tabular}
\caption{We present both a ``restricted" setting with relevant sources to pick between and a ``full" setting in which retrieval includes all sources.
Both VLP\cite{Zhou20VLP} and PICa \cite{yang2021empirical} leverage VinVL\cite{zhang2021vinvl} features. 
CLIP$_{(20)}$ uses VLP to further filter to two sources for QA (Table \ref{tb:Filter}) and is 8pts weaker than the restricted setting.
}
\label{tb:Main}
\end{table}

\subsection {Source Retrieval}

\setlength{\oldcolumnsep}{\columnsep}
\setlength\columnsep{10pt}
\setlength{\oldintextsep}{\intextsep}
\setlength\intextsep{0pt}
\begin{wraptable}[12]{r}{0.55\linewidth}
\small
\begin{tabular}{@{}l@{\hspace{2pt}}lc@{\hspace{5pt}}c@{}}

\toprule
& Query Type & Image & Text \\
\midrule
\multirow{2}{*}{\rotatebox{90}{Restr.}} & \update{BM25} & 25.61 & 43.75 \\
& VLP & \textbf{68.13} & \textbf{69.48} \\
\midrule
\multirow{3}{*}{\rotatebox{90}{Full}} & \update{BM25} & 20.43 & \textbf{28.15} \\
& CLIP$_{(2)}$ & 9.71 & 13.96 \\
& CLIP$_{(20)}$+VLP & \textbf{21.68} & 26.01 \\
\bottomrule
\end{tabular}
\vspace{-5pt}
\caption{Source Retrieval (F1\! $\uparrow$) over \update{$\sim$40 sources (Restr.) or the full corpus. In CLIP$_{(20)}$+VLP, VLP reranks the top 20 sources retrieved.}}
\label{tb:Filter}
\end{wraptable}
Crucial to a complete system design is multimodal source retrieval.
We investigate the effect of retrieval scale (Table \ref{tb:Filter}) and dense versus sparse retrieval approaches. 
For the VLP-based model, sources are selected if its binary classification confidence is above a specified threshold. While the optimal thresholds for different models may vary, for fair comparisons we use 0.2, which is optimal for VLP on the development set. 

\update{VLP achieves $>$68\% F1 given a restricted set of candidates. Indicating that it can model semantic relevance, despite its lack of scalibility. 
In comparison, we use simpler and less expensive approaches when scaling up to the full collection which causes our overall performance to degrade substantially (likely due both to the ambiguity and the weaker underlying document representations). The dense retrieval method suffers from a greater performance drop compared to sparse retrieval. Having VLP rerank the top 20 sources predicted by CLIP doubles F1, which holds promise for a future of large-scale coarse-to-fine retrieval that strikes a better accuracy-efficiency balance. See Appendix D for additional retrieval results.}

\setlength{\columnsep}{\oldcolumnsep}
\setlength{\intextsep}{\oldintextsep}

\subsection{Question Answering}
\begin{table}[t]
\centering
\small
\begin{tabular}{@{}l@{\hspace{2pt}}l@{\hspace{5pt}}c@{\hspace{5pt}}c@{\hspace{5pt}}c@{\hspace{3pt}}c@{\hspace{3pt}}c@{\hspace{3pt}}c@{\hspace{15pt}}c@{}}
\toprule
 & & \multicolumn{6}{c}{Image} & Text \\
 & & \rotatebox{45}{Y/N} & \rotatebox{45}{MC} & \rotatebox{45}{Color} & \rotatebox{45}{Shape} & \rotatebox{45}{Number} & \rotatebox{45}{Other} & \\
& \# samples & 935 & 981 & 228 & 62 & 200 & 1058 & 4067 \\
\midrule
\multirow{4}{*}{\rotatebox{90}{\begin{footnotesize}Fine-Tune\end{footnotesize}}} & VLP (Q-only) & 16.1 & 49.0 & 3.9 & 0.8 & 0.5 & 27.9 & 18.1 \\
& VLP & \textbf{17.2} & 52.9 & 2.8 & 0.0 & 0.5 & \textbf{28.6} & \textbf{50.4} \\
& \VLPi & 11.6 & \textbf{55.3} & \textbf{3.9} & \textbf{2.4} & \textbf{0.5} & 26.3 & ----- \\
& \VLPt & ----- & ----- & ----- & ----- & ----- & ----- & 48.6 \\
\midrule
\multirow{2}{*}{\rotatebox{90}{\begin{footnotesize}Few\end{footnotesize}}} & PICa (Q-only) & 26.7 & 70.1 & 30.8 & 19.0 & 14.2 & 45.3 & 42.8 \\
& PICa & 27.4 & 70.1 & 42.5 & 17.3 & 13.8 & 48.7 & 74.8 \\
\midrule
& Human & \textbf{100} & \textbf{96.8} & \textbf{95.8} & \textbf{94.8} & \textbf{95.0} & \textbf{87.8} & \textbf{94.0} \\
\bottomrule
\end{tabular}
\caption{QA performance breakdown by question categories when presented with oracle sources: $\mathbf{Acc}$ $\uparrow$}
\label{tb:QA_Qcate}
\vspace{-10pt}
\end{table}

\begin{table*}[h]
    \centering
    \begin{tabular}{@{}p{0.30\textwidth}@{\hspace{5pt}}p{0.68\textwidth}@{}}
    Source(s) & Question \textbf{(Q)}, Predicted Answer \textbf{(Pred)}, \& Correct Keywords \textbf{(KW)} \\
    \toprule

\raisebox{-.5\totalheight}{\includegraphics[width=0.49\linewidth, trim=10pt 0 0 25pt, clip]{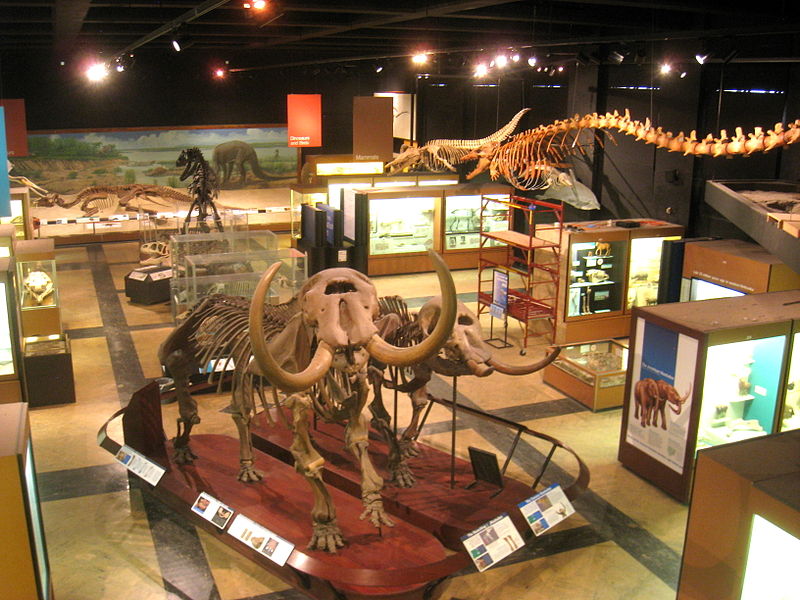} \includegraphics[width=0.49\linewidth, trim=10pt 0 0 0, clip]{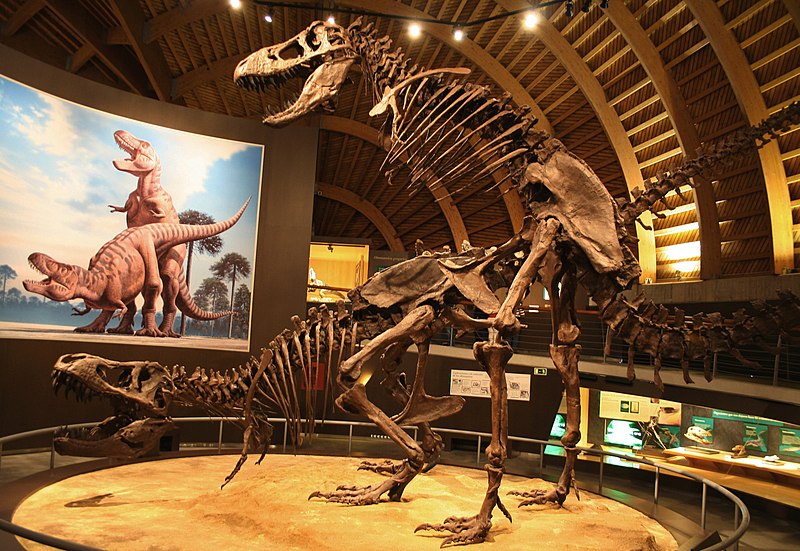}}  & 
\makecell*[{{p{0.95\linewidth}}}]{\small \textbf{Q:} Are the land dinosaurs guarded by rail in both the Display Museum of Natural History in University of Michigan and the Museo Jurassic de Asturias?  \\
\small \textbf{Pred:} No, the land dinosaurs are not guarded by rail. \hfill  
\small\textbf{KW:} Yes  \\
\small \textbf{Notes:} The prediction is wrong but the output sentence is consistent in terms of negation. }\\

\midrule
\raisebox{-.5\totalheight}{\includegraphics[width=\linewidth, trim=0 140pt 0 110pt, clip]{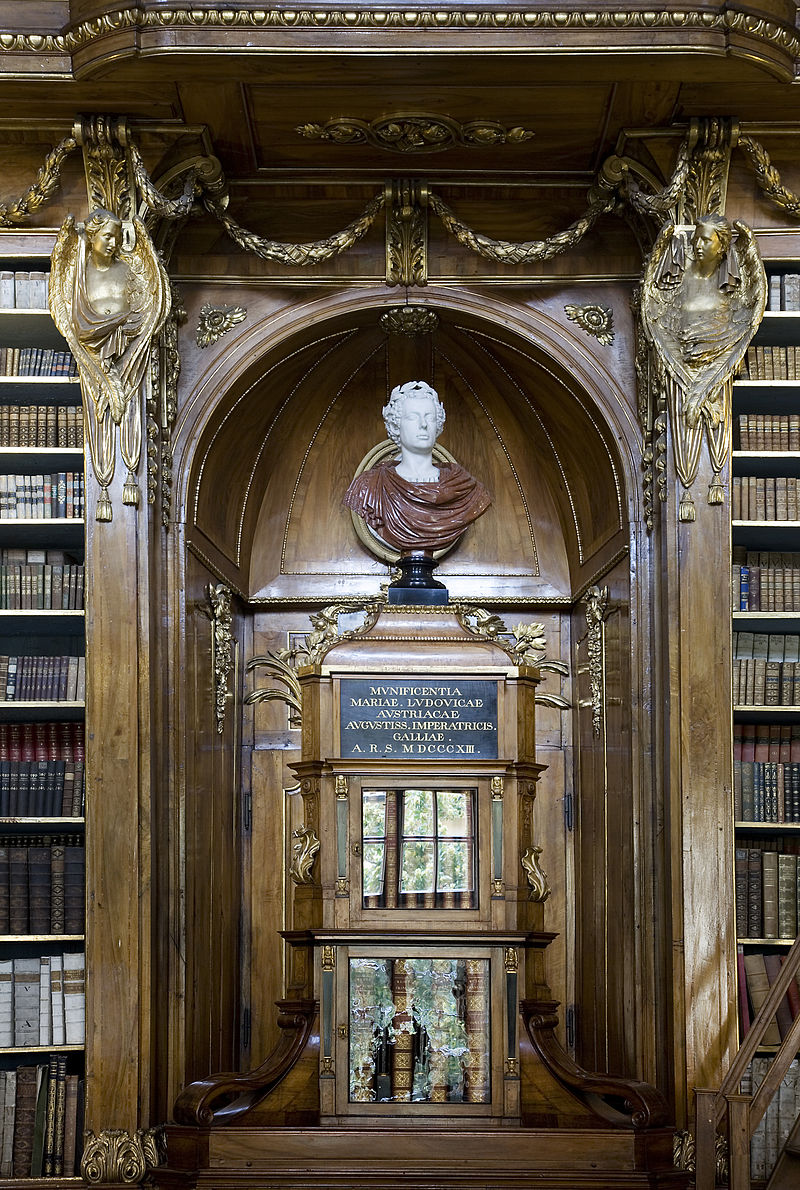}} &
\makecell*[{{p{0.95\linewidth}}}]{\small \textbf{Q:} What is the sculpted bust at the Baroque library, Prague wearing on its head? \\
\small \textbf{Pred:} The sculpted bust at the Baroque library, Prague is wearing a helmet on its head . \hfill
\small \textbf{KW:} A flower wreath \\
\small \textbf{Notes:} The model does not seriously consider the visual information. Comparing to a wreath, helmet is more likely to appear on a head and thereby being a safer choice. }\\
    \end{tabular}

    \begin{tabular}{@{}p{0.57\textwidth}@{\hspace{3pt}}p{0.42\textwidth}@{}}
    \midrule

\makecell*[{{p{0.95\linewidth}}}]{\footnotesize 1. After the heavy 707 quad-jet was introduced in 1958, Boeing addressed the demand for shorter flight lengths from smaller airports. On December 5, 1960, the 727 was launched with 40 orders each from United Airlines and Eastern Air Lines .\\
\footnotesize 2. The first airliner with jet power only was the Nene-powered Vickers VC.1 Viking G-AJPH, which first flew on 6 April 1948.} &
\makecell*[{{p{\linewidth}}}]{\small \textbf{Q:} How many years after the flight of the first jet airliner was the Boeing 727 released ? \\
\small \textbf{Pred:} 727 \\
\small \textbf{KW:} 12 years } \\
\multicolumn{2}{@{}p{0.95\linewidth}@{}}{\small \textbf{Notes:} The model realizes a number is required, but being unable to perform arithmetic, simply copies a number from the snippets.} \\

\midrule
\makecell*[{{p{0.95\linewidth}}}]{\footnotesize 1. Posterior reversible encephalopathy syndrome (PRES), also known as reversible posterior leukoencephalopathy syndrome (RPLS), is a rare condition in which parts of the brain are affected by swelling, usually as a result of an underlying cause. \\
\footnotesize 2. The diagnosis is usually made by brain scan (MRI) on which areas of swelling can be identified. The treatment for PRES is supportive: removal of the cause or causes and treatment of any of the complications, such as anticonvulsants for seizures.} &
\makecell*[{{p{\linewidth}}}]{\small \textbf{Q:} How is the condition also known as reversible posterior leukoencephalopathy syndrome (RPLS) diagnosed? \\
\small \textbf{Pred:} It is diagnosed by swelling , usually as a result of an underlying cause. \\
\small \textbf{KW:} by brain scan (MRI) }\\
\multicolumn{2}{@{}p{0.95\linewidth}@{}}{\small \textbf{Notes:} The selected span from the first source is relevant but does not inform the diagnostic method }\\

\bottomrule
    \end{tabular}
    
    \caption{Common failures (see supplementary for additional predictions) include attempts at extraction or language model hallucinations.}
    \label{tb:qualitative}
\end{table*}

Table \ref{tb:QA_Qcate} provides an accuracy breakdown with respect to question categories. A noticeable pattern is that models are more capable of solving text-based queries than image-based queries. Both VLP and PICa greatly surpasses the question-only baseline and VLP performs favorably against \VLPt, demonstrating reasonable use of sources and the effectiveness of combined training. 

On the other hand, image-based queries pose a much harder challenge. VLP and \VLPi{} are no better than the question-only baseline on image-based queries. While this may be an issue of the sources being ignored, we also attribute this to the fact that the image-based testing samples are intentionally constructed to prevent the success of any superficial correlations that can be drawn from the training set (e.g. the majority answers in each category). We observe a similar issue with PICa. Although PICa consistently outperforms VLP, it does not demonstrate an appropriate utilization of the provided sources, which is especially true on ``Y/N", ``MC", ``Shape", ``Number" and ``Other" question categories. PICa has a surprising amount of knowledge embedded in its parameters, but unlike with text, on images it shows very little improvement from the inclusion of visual sources, as such 
it is still lacking the ability to  explicitly and effectively use the retrieved sources, which might be crucial for further progress towards human accuracy. 

We argue that performance is bottlenecked by the lossy textual representation of images consumed by PICa, thereby calling for concerted effort from both language and vision sides to build a unified representation rather than simply relying on one modality being translated to the other. For future research, we expect to explore whether symbolic or compositional representations in a structured problem space could equip a generative model with skills to perform aggregation beyond simple extraction.

\subsection{Qualitative Analysis}
Finally, we perform a qualitative analysis of the model's failures for both image- and text-based questions. Table \ref{tb:qualitative} includes two image-based and two text-based examples with commentary \update{(additional analysis in Appendix F)}.  Both image questions are clean examples of producing logically consistent and fluent sentences which are incorrect.  The first matches the negation but the answer should have been yes, while in the second, the model runs away with a very logical hallucination (heads wear helmets).

In the text examples, we see a different pattern.  Here the model is more easily able to copy facts from the source texts, but still demonstrates a lack of understanding or reasoning.  In the first example, the model appears to know it is looking for a number, but choosing one via direct copying rather than performing the arithmetic necessary to combine both facts.  In the second case, the model finds a relevant span selection (as is commonly the only thing necessary for text QA tasks), but does not understand that the question is asking about a method of diagnoses versus a symptom.

None of the questions presented here require complex problem-solving skills.  They follow rather simple implication, addition, or visual extraction patterns which are out of reach for current models (uni- or multi-modal).

\section{Conclusion}
\data{} is a new multi-hop, multi-modal question answering challenge for our community. Designed to simulate the heterogeneous information landscape one might expect during a web search, \data{} covers a series of open-domain general visual queries while also forcing models to still reason about text.
Our task requires a system to determine relevant sources, perform aggregation and reasoning. 
We also propose a novel general recipe for evaluation on \data{} which measures both fluency and accuracy. 

Neither the versatile V\&L transformer nor the large-scale text generator present a nearly-there solution. \update{We provide both a restricted and full retrieval setup, to bridge multimodal QA and IR research.} This dataset not only mirrors our everyday experience on the web, but provides a playground for the community to explore important sub-challenges, targeting the creation of a single model for multimodal reasoning, knowledge aggregation, and open-domain visual understanding. 

\data{} aims to facilitate research into constructing a single model which can 1) retrieve relevant documents, and 2) integrate information across a large context window including multiple paragraphs and images, in order to 3) generate fluent natural language answers.

{
    \clearpage
    \small
    \bibliographystyle{ieee_fullname}
    \bibliography{macros,main}
}
\clearpage
\appendix

\section{Data Annotation Details}
\label{app:interfaces}

\paragraph{Qualification HIT}
For quality control, we included a qualification task with 15 hard coded QA pair annotations, some of which obviously violate the annotation guidelines. Annotators had to point out the problematic pairs and explain in what ways they did not follow the instructions. We restricted to crowdworkers located in the US or Canada, with a general requirement of over 1,000 previously approved HITs with at least 95\% approval rate. Additionally, one has to score 80\% or higher on our qualification task before getting access to our main task. We gave workers who achieved 60\% - 80\% at their first attempt a second chance because we believe that workers who had the patience to complete their first attempt were more coachable than others.

\vspace{5pt}
\includegraphics[width=0.85\linewidth]{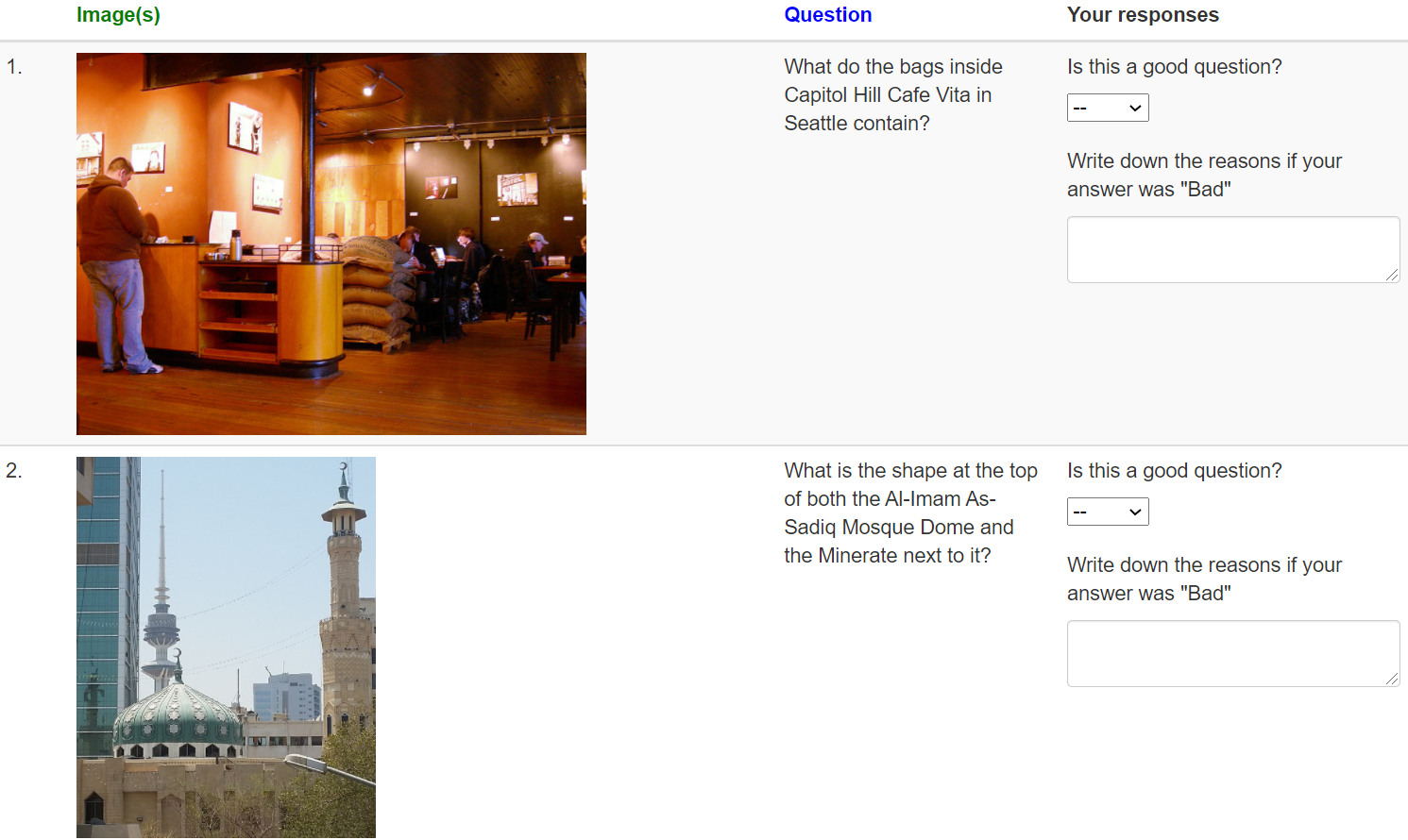}
\vspace{10pt}

\paragraph{Image Filter HIT}
We designed a Filter HIT as a pre-step to obtain groups of related images as prompts for the QA-pair creation task. We present 10 images at a time, which are returned by an Image Search API call using the same search term. Annotators were told to a) select 3 out of the 10 that are distinct but related in some ways, and b) give a label that best summarizes the commonality. After having all these image triples, we paired up triples to form groups 6 according to the cosine similarity between their topic labels. We tuned similarity thresholds to make sure that within each group all images fall under the same topic but still have enough dissimilarity to facilitate both connection-based and comparison-based QA-pair construction.

\paragraph{QA Pair Creation HIT}
The main annotation task (QA-pair creation task) was released batchwise. We spot checked data quality after every batch and sent targeted feedback when we noticed any deviation from our expectations. Workers who constantly failed to follow the guidelines were de-qualified. Crowdsourcing data is challenging in that crowdworders are usually income-driven and will stick to a fixed answer generation pattern once they find it lucrative. To better align the crowdworkers' incentives with our goal, we gave generous bonuses to the annotations that demonstrate out-of-the-box thinking.

\vspace{10pt}
\includegraphics[width=0.9\linewidth]{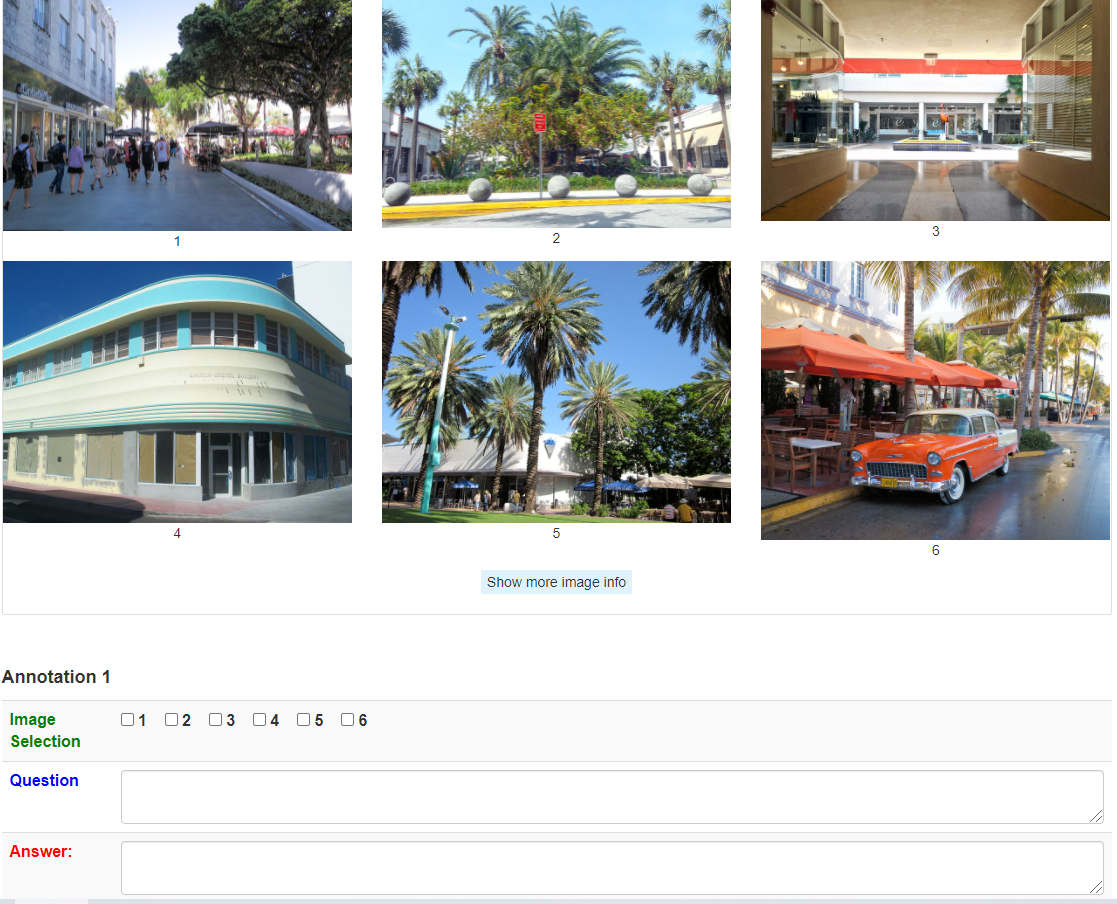}
\vspace{5pt}
\includegraphics[width=0.9\linewidth]{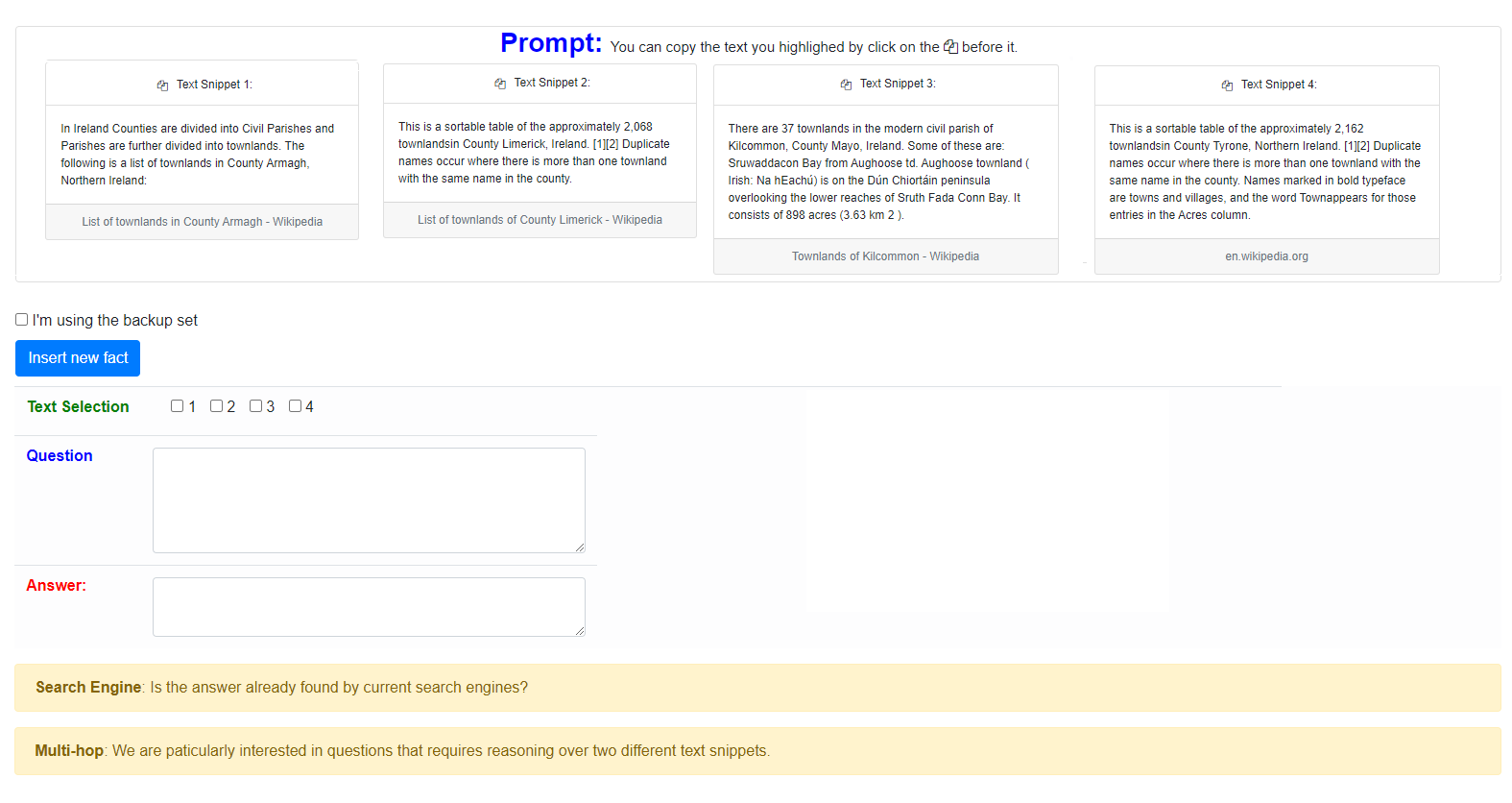}
\vspace{10pt}

\paragraph{QA Pair Validation HIT}

\vspace{5pt}
\includegraphics[width=0.9\linewidth]{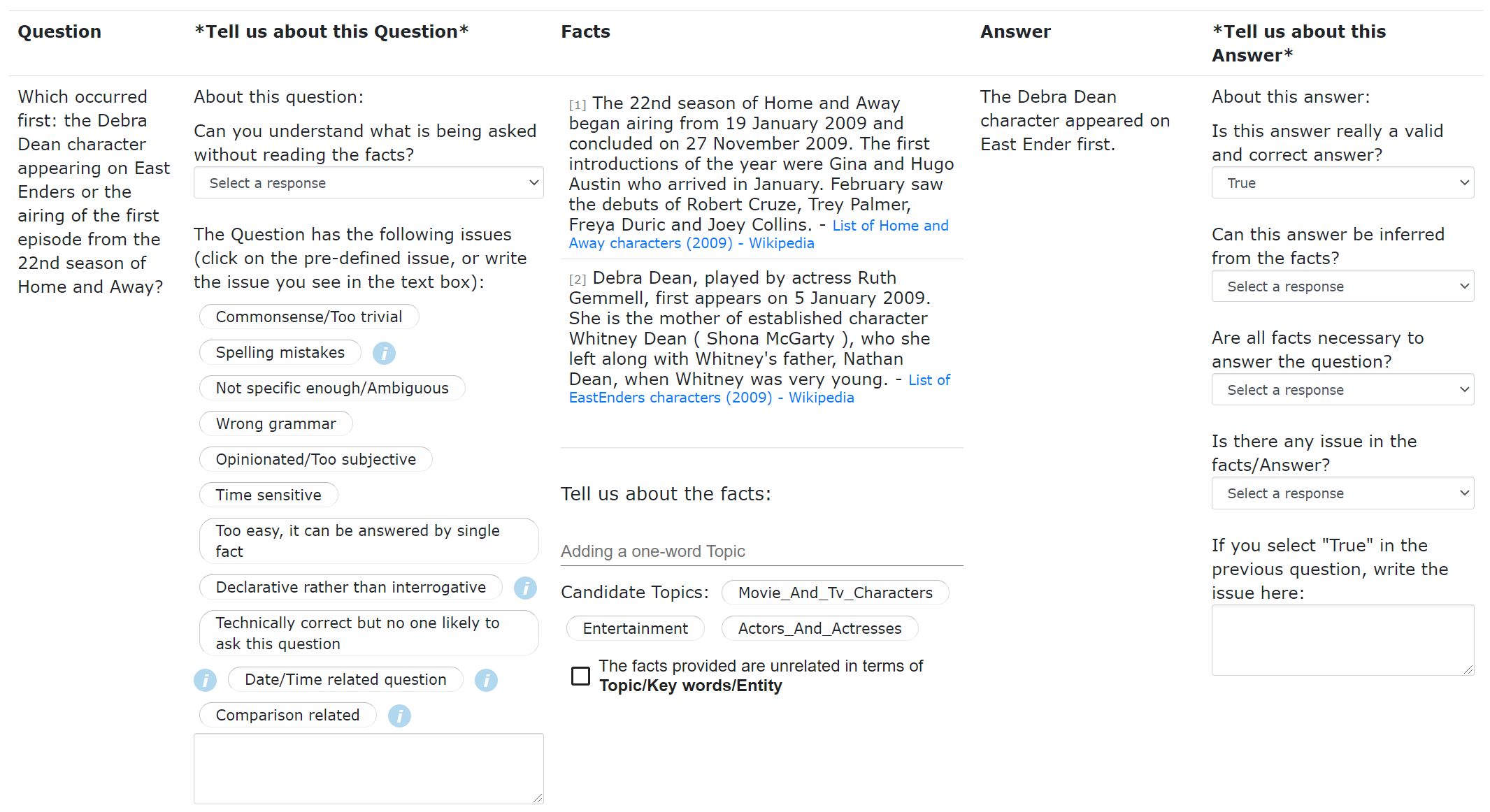}

\paragraph{Multiple Human References Generation HIT}

\vspace{5pt}
\includegraphics[width=0.9\linewidth]{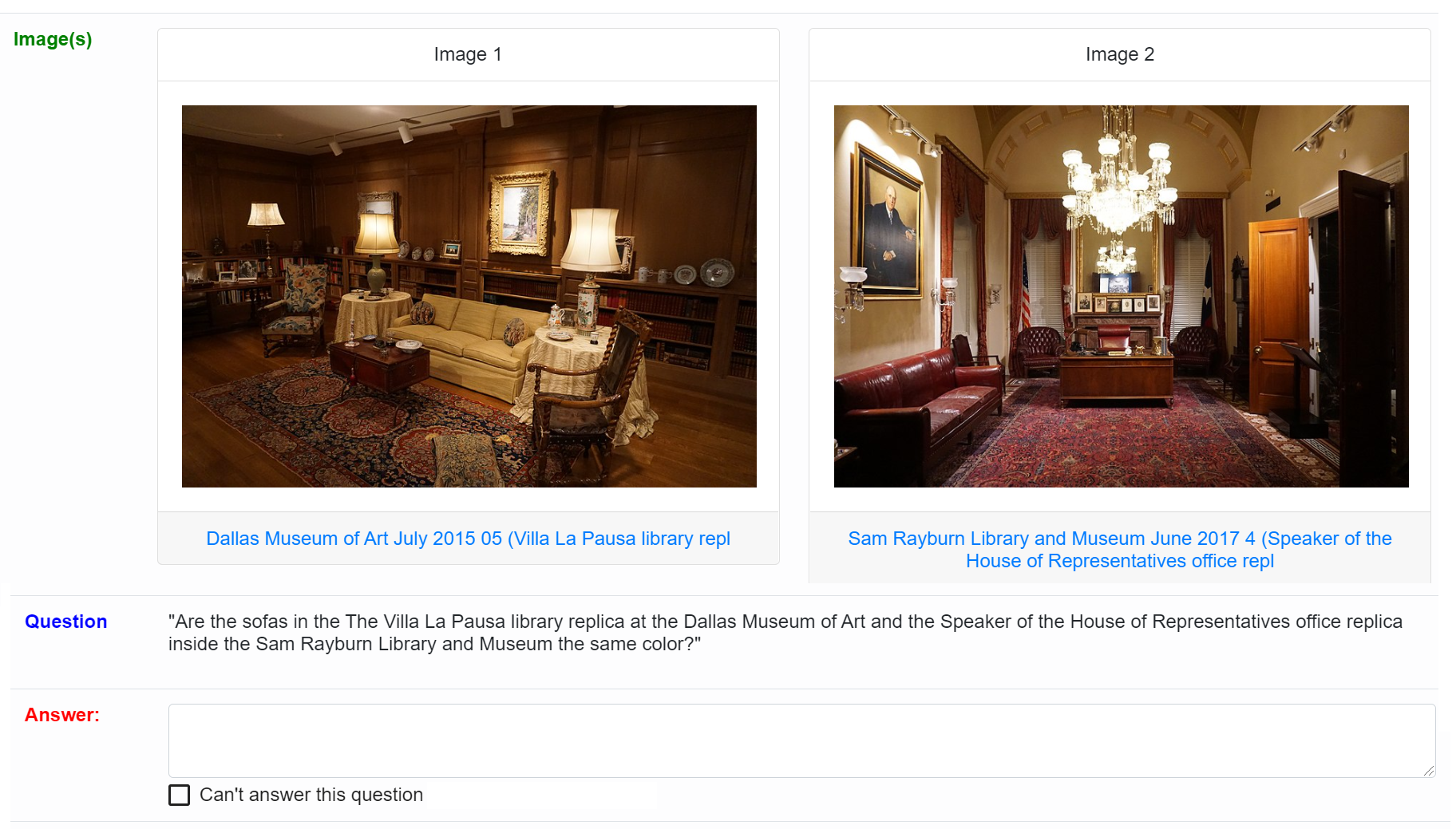}
\includegraphics[width=0.9\linewidth]{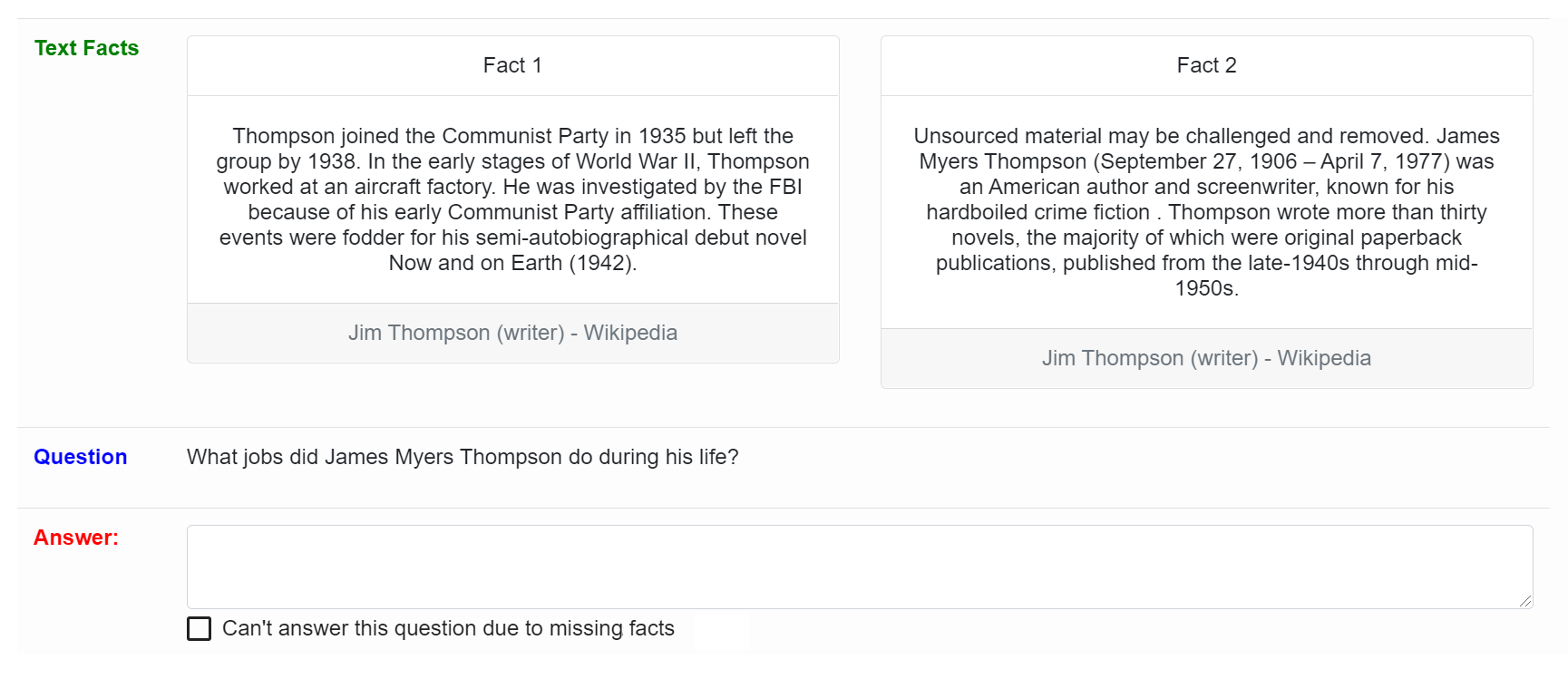}

\section{Visualization of Image Question Prefixes}

\includegraphics[width=0.9\linewidth]{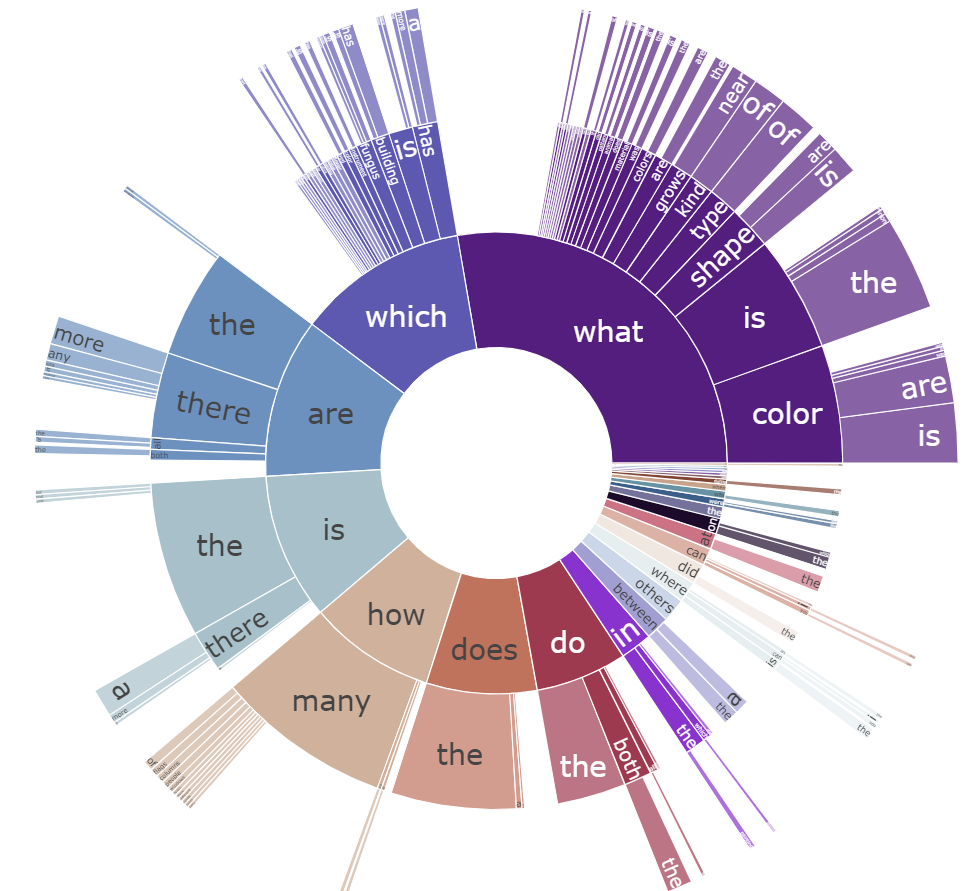}

\section{Classification Based Coverage}

The figure below shows the test set coverage of Top-K training \textbf{keywords} (image-based). All keywords ($>\!$5k) provides only $\sim\!$70\% coverage. The \textbf{full sentence} answers are almost entirely unique, suggesting that classification-based approaches are at a significant disadvantage on WebQA.

\vspace{5pt}
\hspace*{-5pt}\includegraphics[width=0.95\linewidth]{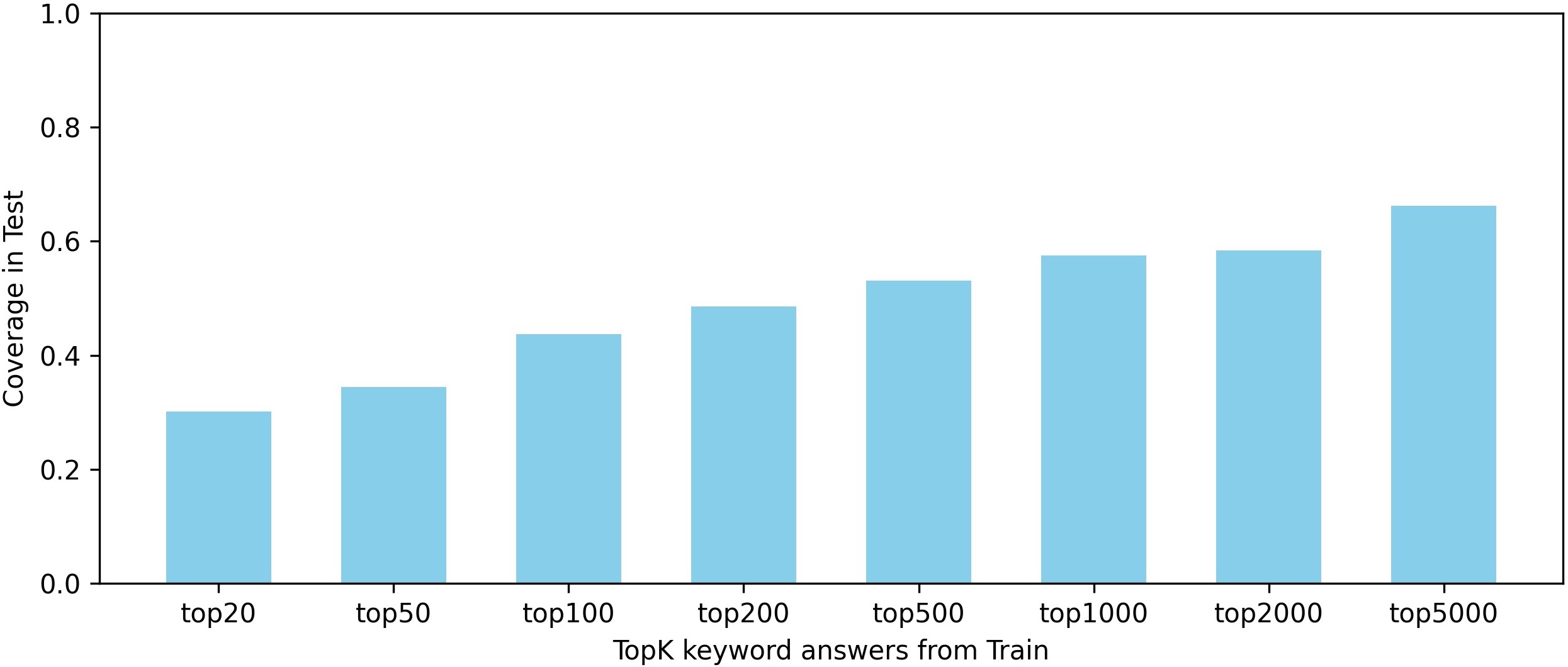}

\section{Additional Results on Full-scale Retrieval}
Assuming known answer modality, CLIP\cite{radford2021learning} achieves 91\%  and 64\% recall rate for image- and text-based queries when 2,000 candidates are retrieved. Without the modality knowledge, the recall rate for image-based queries is zero because the question-image similarities are systematically lower than question-text similarities. Future work may fine-tune dense multimodal retrieval models to close the gap between question-image and question-text similarities.

\includegraphics[width=0.85\linewidth]{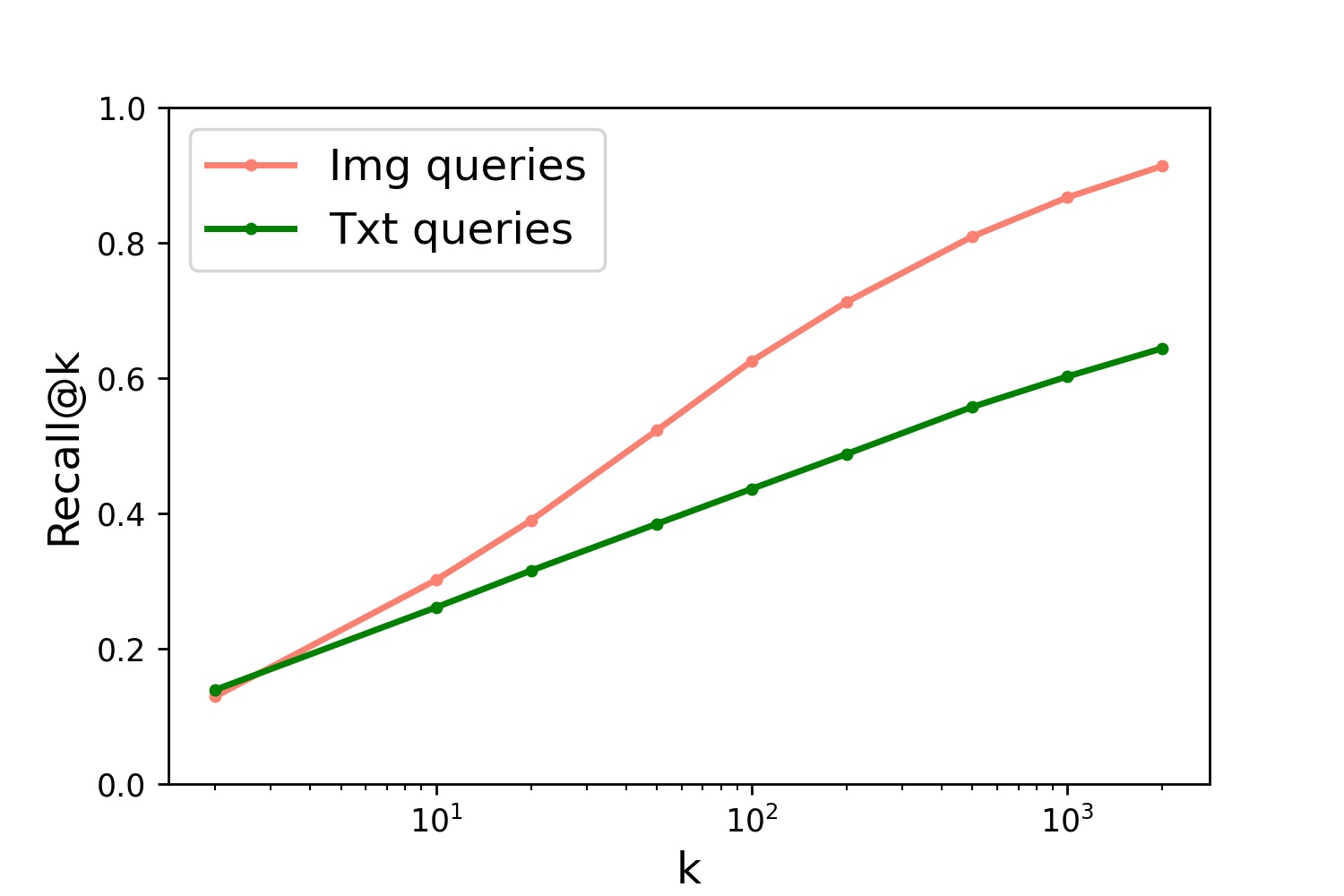}
\label{fig:clip_recall_curve}

\section{Comparing WebQA and recent benchmarks}
\label{app:comparison}

We succinctly contrast WebQA against existing knowledge-aware and multimodal datasets in the main paper. We provide here a more complete  clarification of the new contributions of WebQA over relevant datasets in prior work in terms of data size, modalities and reasoning levels. 

WebQA differs from QAngaroo, HotpotQA, ComplexWebQuestions, HybridQA and NaturalQuestions either in the knowledge-awareness or the involvement of both text and image modalities. OK-VQA, MultiModalQA, ManyModalQA and MIMOQA qualify as both knowledge-seeking and multimodal. Thus we explain them in detail. 

\textbf{OK-VQA}\cite{Marino19okvqa} OK-VQA and our task differ in the role of images. Images in OK-VQA are regarded as part of the query rather than the knowledge source, so source retrieval is not required. However, images in WebQA serve as the knowledge rather than part of the query and can only be processed after retrieval. OK-VQA Topics:

\begin{center}
\includegraphics[width=0.7\linewidth]{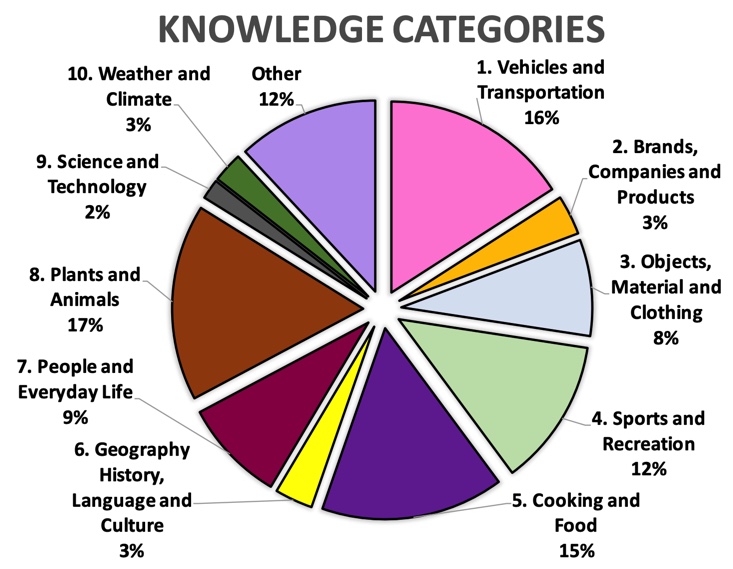}
\end{center}

\textbf{MultiModalQA} \cite{talmor2021multimodalqa} MultiModalQA and WebQA differ in the way qa-pairs were constructed and the answer schema. First, MultiModalQA questions are generated from 
templates. While this facilitates the data generation process, it does not mirror the way real users construct queries. Once the question template is detected, the task reduces to filling in blanks with modality-specific answering mechanisms. This problem-solving manner might not generalize to queries issued by real users where an underlying template is less obvious. In contrast, queries in WebQA are written by annotators, and more structurally diverse. Second, MultiModalQA requires different answer schemas for TextQA, ImageQA and TableQA. TextQA expects a span, “yes” or “no” as an answer. ImageQA expects selection from a fixed answer vocabulary determined by the training set. TableQA expects “yes”, “no”, a table cell, or a summary of more than one table cells via a predicted aggregation operation (i.e. SUM / MEAN / COUNT). We unify the answer schema to be a complete natural language sentence and use an open answer set, so neither span prediction nor classification over a fixed vocabulary suffice. MultiModalQA Topics:

\begin{center}
\includegraphics[width=0.7\linewidth]{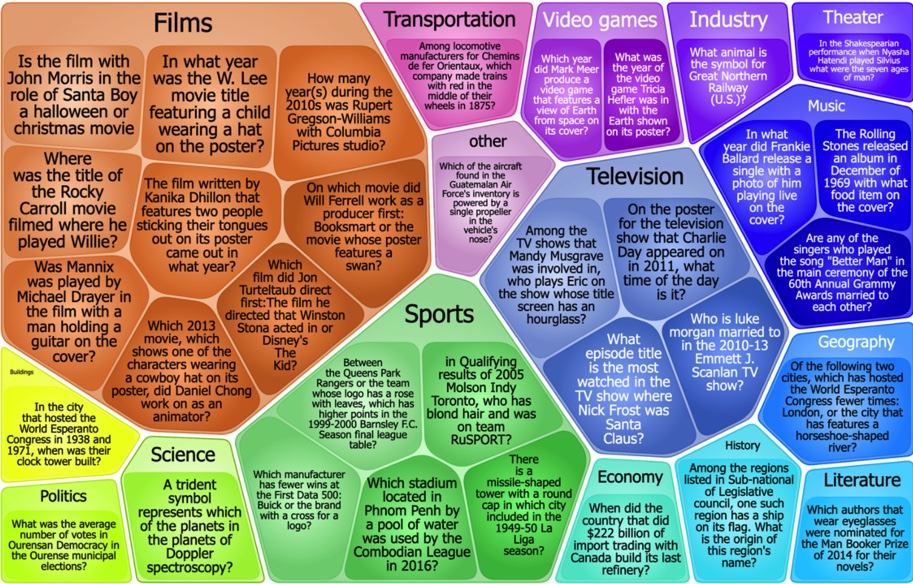}
\end{center}

\textbf{ManyModalQA} \cite{hannan2020manymodalqa} The primary challenge ManyModalQA addresses is the choice of answer modality -- rather than knowledge aggregation or extraction. Our focus is less about distinguishing the answer modality, than about representing world knowledge in a unified space, since mastering the latter may naturally eliminate the need to classify questions into different buckets according to their answer modality. Also, to avoid ambiguity and for easy evaluation, ManyModalQA restricts all answers to be a single word. Therefore, the following question answering is a multiple choice task from [all words in the given context + a pre-defined answer vocabulary]. We argue that multiple choice is an unnatural simplification, because the finite and static answer space imposes a hard limit on the capacity of an answering system, especially when we consider unfamiliar domains, constant shift of world states, and unlimited coverage of the Web. This leads to us formulating WebQA as a free-form generation task, which, although it introduces new challenges for evaluation, better resembles real-world use cases and suits the needs of downstream applications such as voice assistants or conversational agents. Last but not least, ManyModelQA is much smaller than WebQA in size. ManyModalQA Topics:

\begin{center}
    \includegraphics[width=0.8\linewidth]{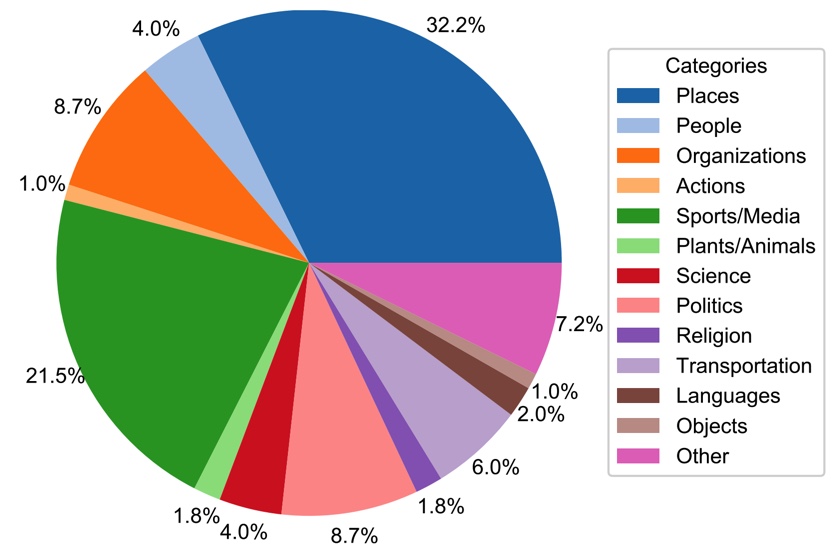}
\end{center}

\textbf{MIMOQA}\cite{singh2021mimoqa} requires selecting a text span from a given context and an image from a set of related images as a multimodal output pair. However, this task formulation does not support queries whose answers should be a digested and summarized version of the given sources instead of a span. WebQA requires further information aggregation and summarization through either numerical or logical reasoning, highlighting the major advantage over MIMOQA in reasoning levels. Plus, WebQA tests natural language generation ability while MIMOQA only requires span prediction and retrieval, both under the classification banner.

\begin{center}
    \includegraphics[width=0.6\linewidth]{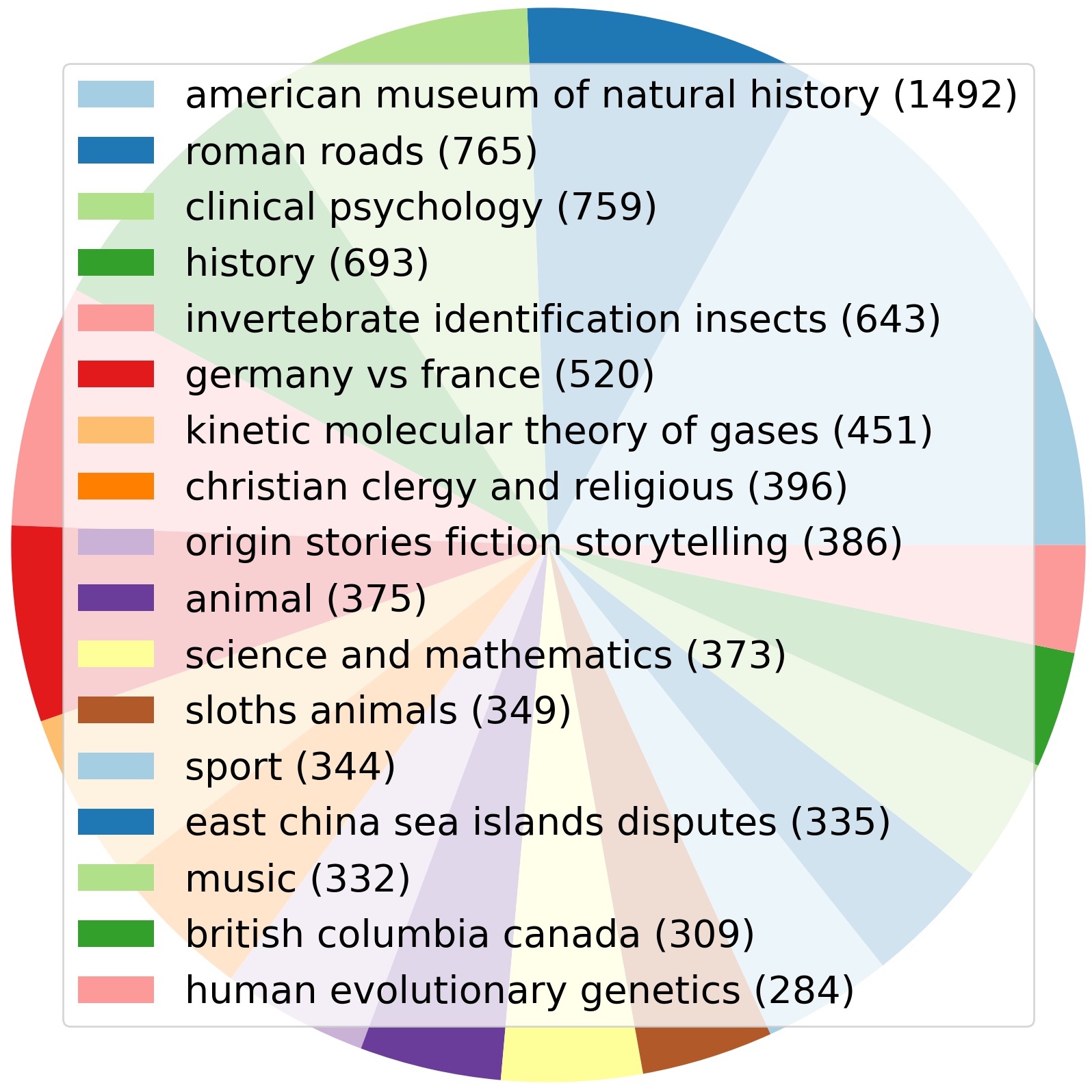}
\end{center}

\noindent
\onecolumn 
\section{Additional Qualitative Analysis}
\begin{small}
    \begin{tabular}{@{}p{0.15\textwidth}p{0.84\textwidth}@{}}
    Source(s) & Question \textbf{(Q)}, Answer Prediction \textbf{(Pred)}, \& Keywords \textbf{(KW)} \\
    \toprule
\raisebox{-.5\totalheight}{\includegraphics[width=0.7\linewidth]{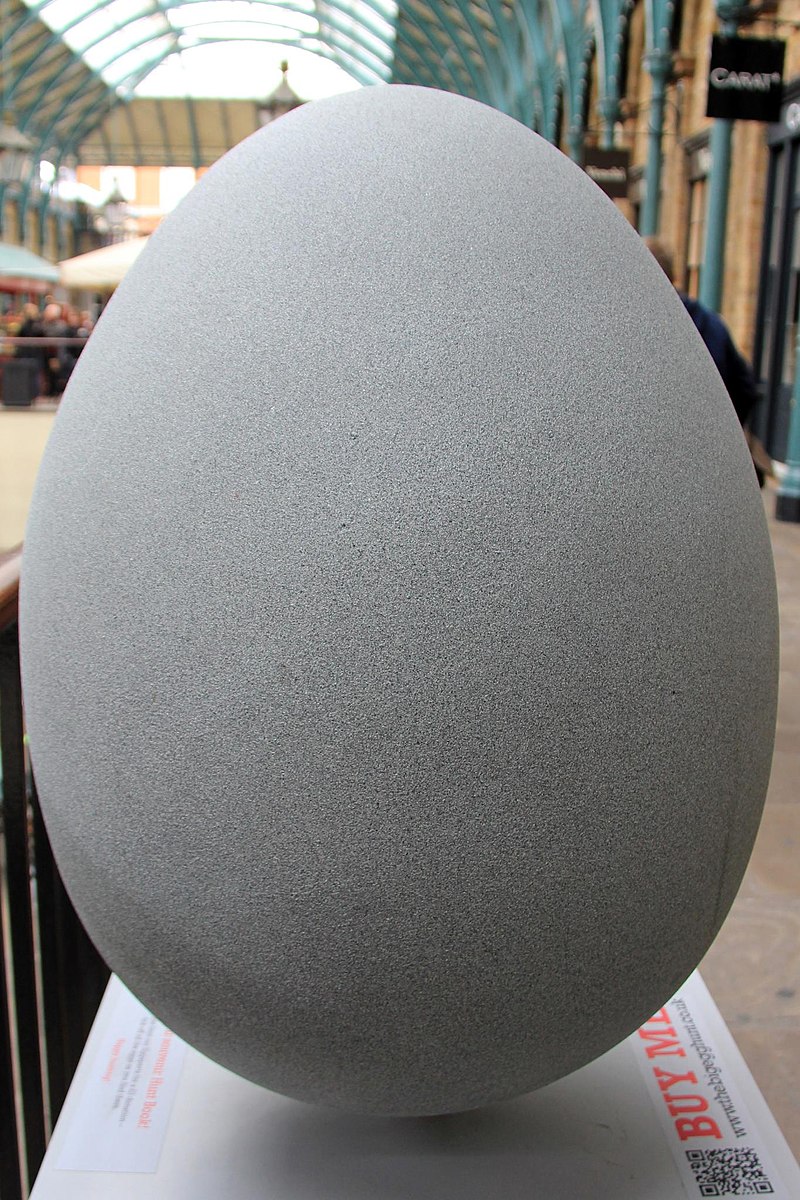}}  &
\makecell*[{{p{\linewidth}}}]{\textbf{Q:} Is the surface of the egg next to the handrail at the Big Egg Hunt in Covent Garden London shiny or dull?  \\
\textbf{Pred:} The surface of the egg next to the handrail at the Big Egg Hunt in Covent Garden London is shiny.  \\
\textbf{KW:} Dull \\
\textbf{Notes:} The model does not have a reasonably large vocabulary for visual properties. It could be the case that  shiny is preferred since it is a more common word in the training set.} \\
\midrule
\raisebox{-.5\totalheight}{\includegraphics[width=\linewidth]{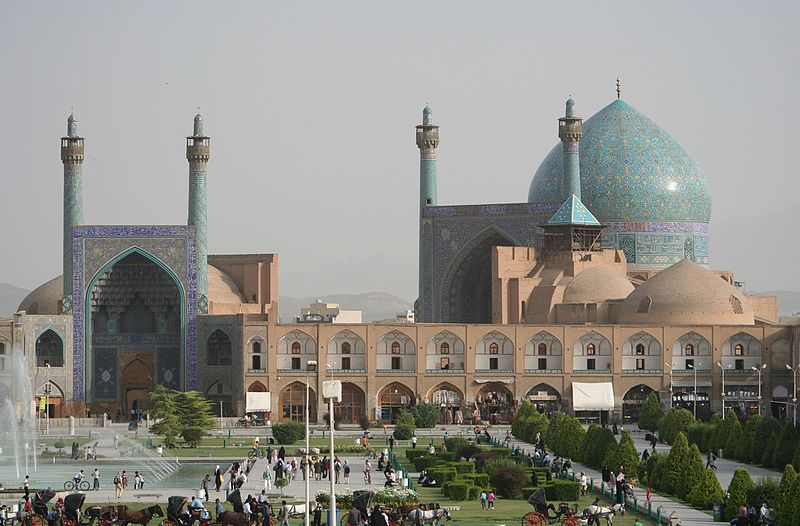}}  &
\makecell*[{{p{\linewidth}}}]{\textbf{Q:} What is the color of the dome of the Isfahan Royal Mosque ?  \\
\textbf{Pred:} The color of the dome of the Isfahan Royal Mosque is white .  \hfill \textbf{KW:} Blue \\
\textbf{Notes:} It can be an  issue of either not looking at the correct region, or not associating the color words with their visual appearances.  But regardless, the superficial pattern matching skills do not work on the adversarial testing samples. }\\

\end{tabular}

    \begin{tabular}{@{}p{0.15\textwidth}p{0.15\textwidth}p{0.67\textwidth}@{}}
    \toprule
\raisebox{-.5\totalheight}{\includegraphics[width=\linewidth]{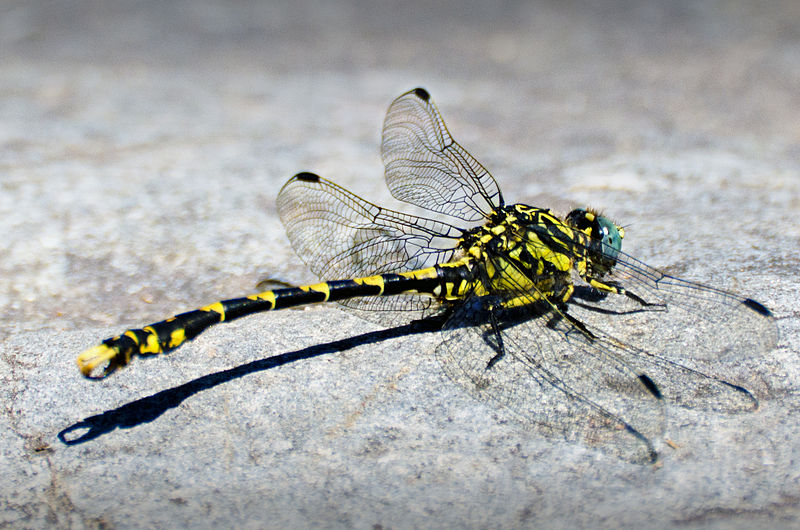}} & \raisebox{-.5\totalheight}{\includegraphics[width=\linewidth]{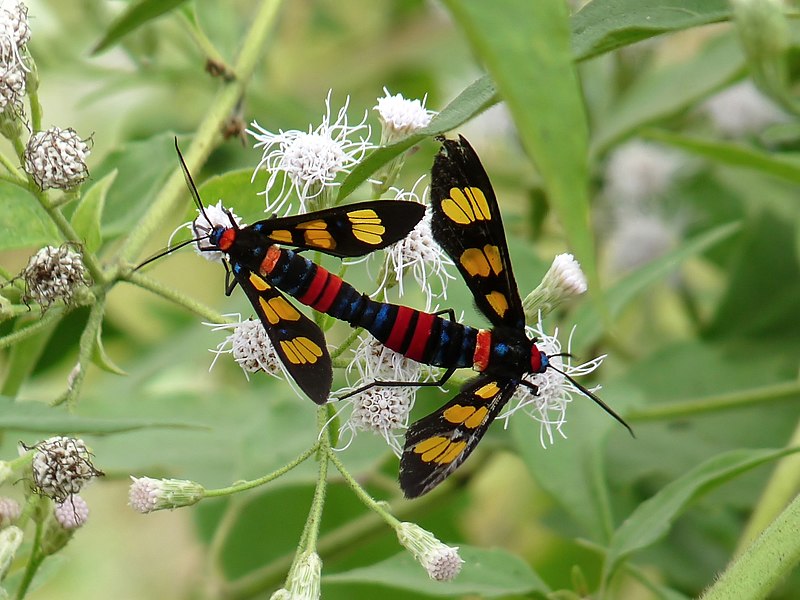}} &
\makecell*[{{p{\linewidth}}}]{\textbf{Q:}What part of the euchromia polymena has the same coloring as the abdomen of the tiger dragonfly ' s abdomen ? \\
\textbf{Pred:} The euchromia polymena has the same coloring as the abdomen of the tiger dragonfly ' s abdomen . \hfill \textbf{KW:} Wings \\
\textbf{Notes:} The model does not understand the question and is treating it as binary.} \\

\midrule
\raisebox{-.5\totalheight}{\includegraphics[width=\linewidth]{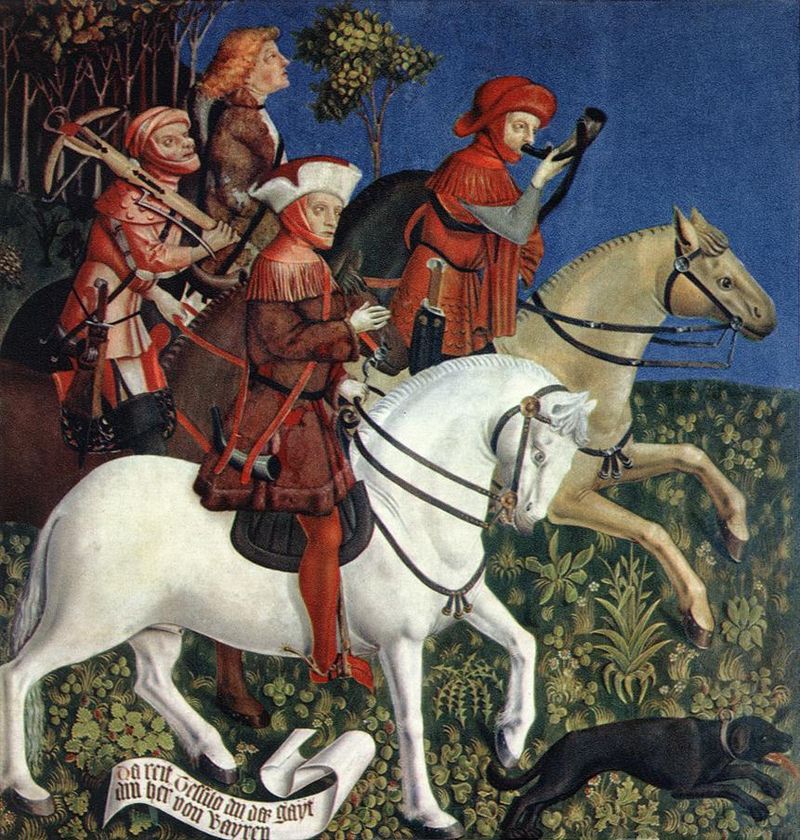}} & \raisebox{-.5\totalheight}{\includegraphics[width=0.8\linewidth]{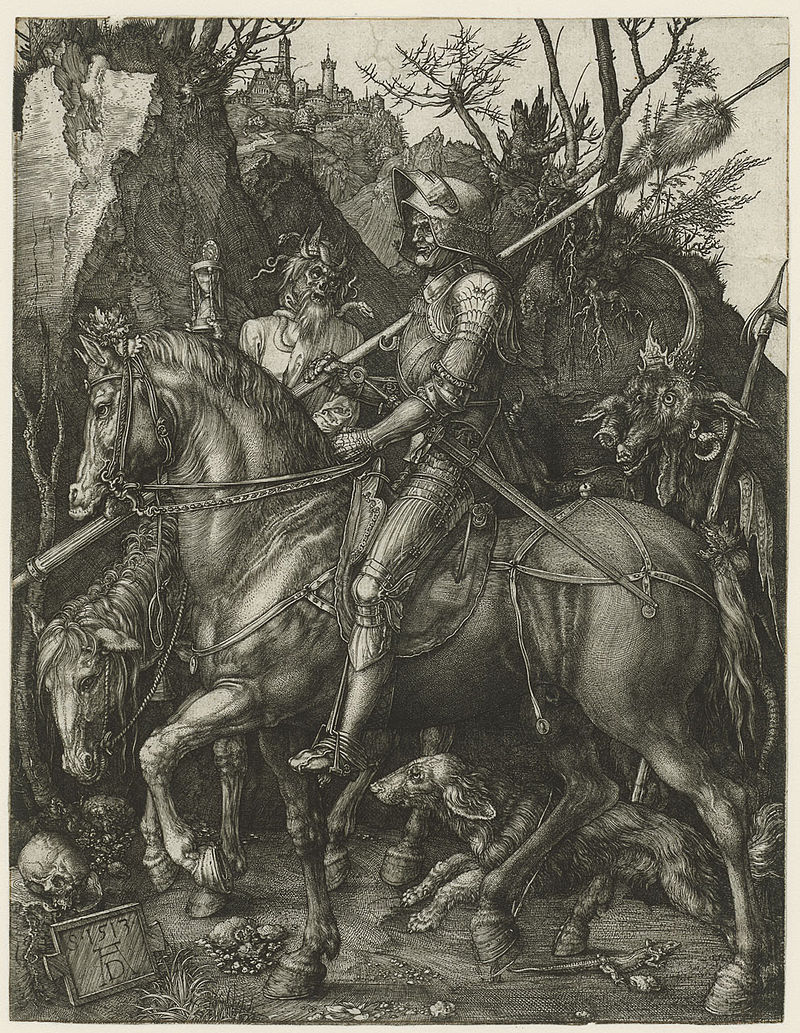}} &
\makecell*[{{p{\linewidth}}}]{\textbf{Q:} What animal is underneath the horses legs in both the paintings Knight , Death , and the Devil and Prince Tassilo Rides to Hunting ? \\
\textbf{Pred:} A dog is underneath the horses legs in both the Knight , Death , and the Devil and Prince Tassilo Rides to Hunting. \hfill \textbf{KW:} A dog \\
\textbf{Notes:} The model predicts correctly, probably due to precise object detection. }\\

    \end{tabular}

    \begin{tabular}{@{}p{0.65\textwidth}p{0.35\textwidth}@{}}
    \toprule

\makecell*[{{p{\linewidth}}}]{\footnotesize 1. Coinage was used in the Ptolemaic Kingdom during the last dynasty of Egypt and, briefly, during Roman rule of Egypt . Ptolemaic coinage was struck in Phoenician weight, also known as Ptolemaic weight (about 14.2 grams). This standard, which was not used elsewhere in the Hellenistic world, was smaller than the dominant Attic weight.\\[5pt]
\footnotesize 2. All the male rulers of the dynasty took the name Ptolemy, while queens regnant were all called Cleopatra, Arsinoe or Berenice. The most famous member of the line was the last queen, Cleopatra VII, known for her role in the Roman political battles between Julius Caesar and Pompey, and later between Octavian and Mark Antony. } &
\makecell*[{{p{\linewidth}}}]{\textbf{Q:} What type of currency was used during Cleopatra VII ' s reign ? \\
\textbf{Pred:} Ptolemaic coinage . \\
\textbf{KW:} Ptolemaic coinage }\\
\multicolumn{2}{@{}p{\linewidth}}{\textbf{Notes:} The model picks up the correct entity} \\
\bottomrule
    \end{tabular}

\end{small}

\twocolumn

\clearpage
\clearpage
\section{Datasheet for \data{}}

\subsection{Motivation}
\paragraph{For what purpose was the dataset created?}

\data{} was created to drive the research progress in multihop, multimodal question answering, which would bridge the gap between the natural language and vision community. 

\paragraph{Who created the dataset (e.g., which team, research group) and on behalf of which entity (e.g., company, institution, organization)?} 

The initial version of \data{} was created by Yingshan Chang and Yonatan Bisk on behalf of Language Technology Institute, Carnegie Mellon University, and Mridu Narang at Microsoft Bing. 

\paragraph{Who funded the creation of the dataset? } 
Microsoft Research and Bing provided the funds for crowdsourcing and web crawling. 

\subsection{Composition} 
\paragraph{What do the instances that comprise the dataset represent (e.g., documents, photos, people, countries)?} 
Each instance is a tuple of (Knowledge Sources, Question, Answer), where a knowledge source can be either an image assisted by a caption, or a snippet. Questions and Answers are in textual form.

\paragraph{How many instances are there in total (of each type, if appropriate)?} 
\data{} is structured as having answers that can be found either via image search or general web (text) search. So there are two folds of data, containing 22,423 image-based queries and 24,343 text-based queries, respectively. There are 600K images crawled from  Wikipedia and 750K snippets crawled from the general Web (mostly from Wikipedia) serving as potential knowledge sources.

\paragraph{Does the dataset contain all possible instances or is it a sample (not necessarily random) of instances from a larger set?}
\data{} is a sample of instances. It is presumably intended to be a random sample of instances representing what one might encounter during a real web search experience. Manual efforts were put in to ensure reasonable coverage and diversity. Only qualitative tests were run to show the inclusiveness.

\paragraph{What data does each instance consist of?} 
Each data instance consists of text and images.

\paragraph{Is there a label or target associated with each instance?} The answer component is regarded as the target. Each instance is associated with one human-written answer in the format of a complete natural language sentence. Additionally, each instance in the testing set has multiple (3-6) full sentence answers as well as a keyword answer annotated by humans, which is supposed to be a succinct rephrasing of the corresponding long-form answer.

\paragraph{Is any information missing from individual instances?}
Everything is included.

\paragraph{Are relationships between individual instances made explicit (e.g., users’ movie ratings, social network links)?} 
There are no relationships between instances except for the fact that multiple instances may share knowledge sources.

\paragraph{Are there recommended data splits (e.g., training, development/validation, testing)?}
The dataset comes with specified train/dev/test splits. The split on the text-based fold was determined randomly while the test split on the image-based fold was adversarialy selected to prevent spurious shortcut learning from inflating the metrics.

\paragraph{Are there any errors, sources of noise, or redundancies in the dataset?}
Erroneous instances were pruned during the validation process after the initial collection, where we had human annotators report mistakes and inconsistency. The released version is clean. 

\paragraph{Is the dataset self-contained, or does it link to or otherwise rely on external resources (e.g., websites, tweets, other datasets)?}
No. All the information crawled from the Web was downloaded and fixed when the dataset was constructed.

\paragraph{Does the dataset contain data that might be considered confidential (e.g., data that is protected by legal privilege or by doctorpatient confidentiality, data that includes the content of individuals’ non-public communications)?} 
No. All data was derived from crowdsourcing and publicly available content on the web.

\paragraph{Does the dataset contain data that, if viewed directly, might be offensive, insulting, threatening, or might otherwise cause anxiety?} No, data was specifically pulled from known vetted resources (e.g. Wikipedia / Wikimedia).

\paragraph{Does the dataset relate to people?} No

\subsection{Collection Process}
\paragraph{How was the data associated with each instance acquired?}
The questions and answers were curated by crowdsourcing. The knowledge sources were mined from the web that were directly observable.

\paragraph{What mechanisms or procedures were used to collect the data (e.g., hardware apparatus or sensor, manual human curation, software program, software API)?}
Crowdsourcing relied on the Amazon Mechanical Turk platform. Web crawling was assisted by Bing Visual Search and Wikipedia APIs.

\paragraph{If the dataset is a sample from a larger set, what was the sampling strategy (e.g., deterministic, probabilistic with specific sampling probabilities)?}
All question-answer pairs were human-curated. Knowledge sources for each sample are determined by their relevance to the question-answer pair.

\paragraph{Who was involved in the data collection process (e.g., students, crowdworkers, contractors) and how were they compensated (e.g., how much were crowdworkers paid)?}
Crowdworkers are paid with an average hourly wage above \$13. 

\paragraph{Over what timeframe was the data collected?} 
\data{} was collected and validated from Oct 2020 to Aug 2021. 

\paragraph{Were any ethical review processes conducted (e.g., by an institutional review board)?}
No

\paragraph{Does the dataset relate to people?} No

\subsection{Preprocessing/Cleaning/Labeling}
\paragraph{Was any preprocessing/cleaning/labeling of the data done (e.g., discretization or bucketing, tokenization, part-of-speech tagging, SIFT feature extraction, removal of instances, processing of missing values)?}
After the initial collection, each sample was validated by 2 or 3 crowdworkers. Problematic samples were discarded. Testing samples with low human agreements were discarded. Besides, each sample in the image-based fold was assigned a question category label produced by a text analysis algorithm.

\paragraph{Was the “raw” data saved in addition to the preprocessed/cleaned/labeled data (e.g., to support unanticipated future uses)?}
The raw unprocessed data (consisting of crowdsourcing output, history versions of unpruned dataset) is saved.

\paragraph{Is the software used to preprocess/clean/label the instances available?}
While a script running a sequence of commands is not available, all codes used to process the data is open source on Github.

\subsection{Uses}
\paragraph{Has the dataset been used for any tasks already?} 
The dataset was introduced in the paper \data{}: Multihop and Multimodal QA. 

\paragraph{Is there a repository that links to any or all papers or systems that use the dataset?}
Papers using this dataset will be listed in \url{https://webqna.github.io/} or linked from the EvalAI leaderboard.

\paragraph{What (other) tasks could the dataset be used for?} 
\data{} can be used for modelling works in the areas of knowledge retrieval, multimodal reasoning and open-domain question answering.

\paragraph{Is there anything about the composition of the dataset or the way it was collected and preprocessed/cleaned/labeled that might impact future uses?}
No. There is minimal known risks for harm.

\paragraph{Are there tasks for which the dataset should not be used?}
Not to our knowledge

\subsection{Distribution}
\paragraph{Will the dataset be distributed to third parties outside of the entity (e.g., company, institution, organization) on behalf of which the dataset was created?}
Yes. \data{} will be made publicly available.

\paragraph{How will the dataset will be distributed (e.g., tarball on website, API, GitHub)?}
See \url{https://webqna.github.io/} for downloading instructions.

\paragraph{When will the dataset be distributed?}
\data{} will be released to the public in Sep 2021.

\paragraph{Will the dataset be distributed under a copyright or other intellectual property (IP) license, and/or under applicable terms of use (ToU)?}
The crawled data copyright belongs to the websites that the data originally appeared in (e.g. Wikimedia Foundation). \data{} will be distributed under freely to academic researchers upon request.

\paragraph{Have any third parties imposed IP-based or other restrictions on the data associated with the instances?}
No

\paragraph{Do any export controls or other regulatory restrictions apply to
the dataset or to individual instances?} No

\subsection{Maintenance}
\paragraph{Who is supporting/hosting/maintaining the dataset?}
\data{} is supported and maintained by Language Technologies Institute @CMU and Microsoft Research, and the leaderboard is hosted on \href{https://eval.ai/}{EvalAI}.

\paragraph{How can the owner/curator/manager of the dataset be contacted (e.g., email address)?}
\{yingshac, ybisk\}@cs.cmu.edu

\definecolor{fuchsia}{rgb}{1.0, 0.0, 1.0}
\paragraph{Is there an erratum?}
All changes to the dataset will be announced on \url{https://webqna.github.io/}

\paragraph{Will the dataset be updated (e.g., to correct labeling errors, add new instances, delete instances)?}
All updates (if necessary) will be posted on \url{https://webqna.github.io/}

\paragraph{If the dataset relates to people, are there applicable limits on the retention of the data associated with the instances (e.g., were individuals in question told that their data would be retained for a fixed period of time and then deleted)?}
\data{} is not related to people.

\paragraph{Will older versions of the dataset continue to be supported/hosted/maintained?}
All changes to the dataset will be announced on \url{https://webqna.github.io/}. Outdated versions will be kept around for consistency. 

\paragraph{If others want to extend/augment/build on/contribute to the dataset, is there a mechanism for them to do so?}
Any extension/augmentation by an external party should be made after contacting the original authors.

\clearpage



\end{document}